\documentclass[lettersize,journal]{IEEEtran}
\usepackage{amsmath,amsfonts}
\usepackage{algorithmic}
\usepackage{algorithm}
\usepackage{array}
\usepackage[caption=false,font=normalsize,labelfont=sf,textfont=sf]{subfig}
\usepackage{textcomp}
\usepackage{stfloats}
\usepackage{url}
\usepackage{verbatim}
\usepackage{graphicx}
\usepackage{cite}

\usepackage{amssymb}
\usepackage{bm}
\usepackage{amsthm,amsmath,amssymb}
\usepackage{mathrsfs}
\usepackage{multirow}
\usepackage{makecell}
\usepackage{enumerate}
\usepackage{color}
\usepackage{stfloats}
\usepackage{comment}
\usepackage{setspace}

\begin{document}

\title{Efficient Image-Text Retrieval via Keyword-Guided Pre-Screening}

\author{Min~Cao,
        Yang~Bai, 
        Jingyao~Wang, 
        Ziqiang~Cao$^*$, 
        Liqiang~Nie,~\IEEEmembership{Senior Member,~IEEE} 
        and~Min~Zhang,~\IEEEmembership{Member,~IEEE} 

\thanks{Ziqiang Cao is the corresponding author (email: zqcao@suda.edu.cn).}%<-this % stops a space
\thanks{
Min Cao, Yang Bai, Jingyao Wang, Ziqiang Cao and Min Zhang are with School of Computer Science and Technology, Soochow University, Suzhou 215006, China (email: mcao@suda.edu.cn).}
 \thanks{
Liqiang Nie and Min Zhang are Institute of Computing and Intelligence, Harbin Institute of Technology (Shenzhen), China.}
% <-this % stops a space
}    

\markboth{Journal of \LaTeX\ Class Files,~Vol.~14, No.~8, August~2021}%
{Shell \MakeLowercase{\textit{et al.}}: A Sample Article Using IEEEtran.cls for IEEE Journals}


\maketitle

\begin{abstract}
Under the flourishing development in performance, current image-text retrieval methods suffer from $N$-related time complexity, which hinders their application in practice.
Targeting at efficiency improvement, this paper presents a simple and effective keyword-guided pre-screening framework for the image-text retrieval.
%We exclude large amounts of irrelevant galleries before the common retrieval network, for which the image and text data are discretized into the keywords and the keyword retrieval across modalities is executed. 
Specifically, we convert the image and text data into the keywords and perform the keyword matching across modalities to exclude a large number of irrelevant gallery samples prior to the retrieval network.
For the keyword prediction, we transfer it into a multi-label classification problem and propose a multi-task learning scheme by appending the multi-label classifiers to the image-text retrieval network to achieve a lightweight and high-performance keyword prediction.
%Specifically, we propose a multi-task learning scheme by appending the multi-label classifiers to the image-text retrieval network to achieve a lightweight and high-performing classification.
For the keyword matching, we introduce the inverted index in the search engine and create a win-win situation on both time and space complexities for the pre-screening.
Extensive experiments on two widely-used datasets, \emph{i.e.}, Flickr30K and MS-COCO, verify the effectiveness of the proposed framework.
The proposed framework equipped with only two embedding layers achieves $O(1)$ querying time complexity, while improving the retrieval efficiency and keeping its performance, when applied prior to the common image-text retrieval methods.
Our code will be released.
\end{abstract}


\section{Introduction}

Recent years have witnessed that cross-modal image-text retrieval has been gradually becoming one of the mainstream research topics in the fields of multimedia computing and information retrieval~\cite{cao2022image,li2021align,radford2021learning,chen2019uniter}. 
It aims to retrieve the gallery samples in one modality from a large-scale repository with a given query sample in another, whereby the cross-modal alignment is well-established to estimate the pairwise similarity between the query and each gallery sample. 
%and modeling the alignment between two modalities, thus achieving image-to-text retrieval and text-to-image retrieval.
Specifically, taking a text as the query to retrieve its corresponding images is called text-to-image retrieval, and vice versa.
Image-text retrieval is challenging due to the heterogeneity and semantic gap between these two modalities.

\begin{figure}[t]
\begin{center}
   \includegraphics[width=1.0\linewidth, height=0.74\linewidth]{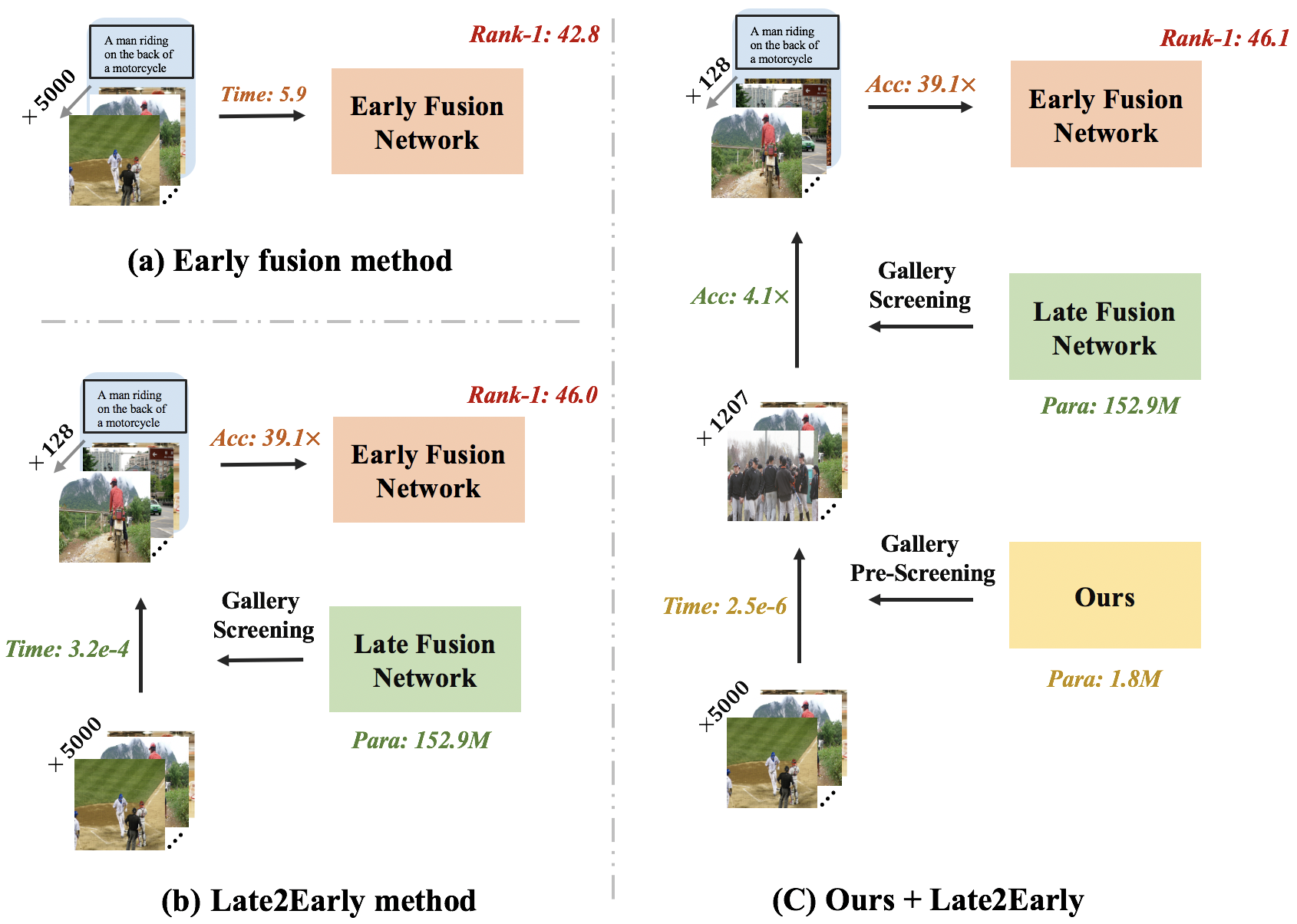}
\end{center}
   \caption{Illustration of various image-text retrieval methods. Rank-1 denotes the expectation of correct match at the 1-th in the ranking list; Time represents the online running time (in minutes); Acc is the abbreviation for acceleration; Para denotes the number of modal parameters (in millions). These results are computed based on the early fusion network ViLT~\protect\cite{kim2021vilt} and the late fusion one ALBEF$_{0}$~\protect\cite{li2021align} on MS-COCO. In the proposed framework, we adopt the classification by a multi-task learning on the ALBEF. 
   It is worth noting that the running time varies across platforms and here we lay stress on the comparison among the running times.}
\label{fig1}
\end{figure}

%Nonetheless, there is still a long thread of work [x] on the image-text retrieval. 
%These works have been conducted to reduce discrepancy between image and text and achieve information alignment between them.
Broadly speaking, the studies on image-text retrieval are in two variants: late and early fusion. 
%image/text processing branches in which the feature information in each model is firstly explored independently, and interaction module in which the resulting feature in each model has further interacted with each other.
In particular, the former~\cite{liu2017learning,wang2018learning,chen2021learning,radford2021learning,andonian2022robust,lu2022cots} emphasizes the image and text feature encodings separately, and then utilizes a simple inner product between the image and text features to estimate the similarities. 
%than on the interaction module, as a result of which the pairwise similarities are computed by only a simple inner product interaction during inference.
Its antithesis is the early fusion methods~\cite{liu2020graph,kim2021vilt,li2020oscar,zhang2021vinvl}, paying more attention to designing complex interaction modules for deeply fusing the image and text features.
%the interaction module, in which the feature in each model is further updated by fusing the information across modalities.
%In recent years, with the success of Transformer architecture~\cite{vaswani2017attention}, large-scale pretraining~\cite{li2020oscar,zhang2021vinvl} gradually becomes a central paradigm in vision-and-language (V+L) tasks including the image-text retrieval task. In these pretraining method, the cross-modal information is usually fused early and freely by a single-stream transformer-like network, and are essentially classified as the early-fused methods.
Thanks to the deep fusion of cross-modal information, the early fusion methods have been the leading paradigm in boosting the search performance.
%of the image-text retrieval task. 
It, nevertheless, suffers from a huge gap between the laboratory experiments and the real-world applications since we need to online compute the feature representations in the interaction module quadratically by traversing through all image-text pairs, leading to the prohibitive reference cost. 
%Particularly, as the current state-of-the-art paradigm, the cross-modal pretraining method is usually characterized by the bulky network architecture, aggravating the problem of high time complexity.
%The accuracy of the early-fused method usually outperforms the later-fused one due to such information fusion. 
%The early-fused method has a performance advantage compared to the later-fused method due to such information fusion. 
%Such information fusion is beneficial to achieve semantic alignment between modalities, thus the early-fused method has an advantage in terms of performance compared with the later-fused method.
%which has been verified in various image-text retrieval studies [x].
%However, we need to compute online the feature by the early-fused method quadratically, it is not economically practical to apply it in reality.
Taking the text-to-image retrieval task as an example, we show a simple flowchart of the early fusion method with its online running time in minutes on the MS-COCO dataset~\cite{lin2014microsoft} in Fig.~\ref{fig1} (a). 
% with the fps and Rank-1 by ViLT~\cite{kim2021vilt} on the Flickr30K dataset~\cite{plummer2015flickr30k} 
To address the low efficiency problem, some researchers~\cite{miech2021thinking,sun2021lightningdot,li2021align,liu2021inflate} propose to first screen the gallery samples by a fast late fusion technique and then retrieve the remaining samples by a slow early fusion technique (Late2Early for short), thus yielding a win-win on accuracy and efficiency, as shown in Fig.~\ref{fig1} (b).
However, an online search operation with $O(N)$ time complexity for the gallery screening is needed in the Late2Early method, still restricting its application in reality where there are usually millions of galleries with a tremendous number $N$.

\begin{figure}[t]
\begin{center}
   \includegraphics[width=1.0\linewidth, height=0.47\linewidth]{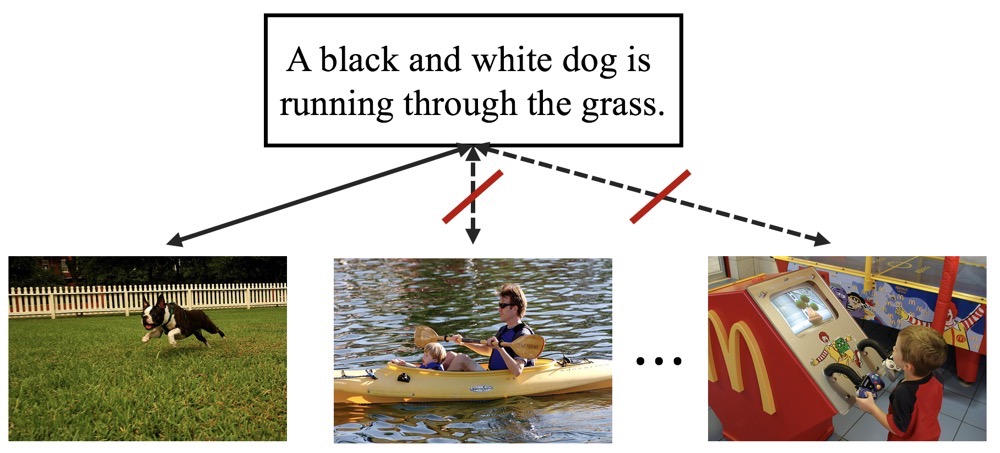}
\end{center}
   \caption{An exposition of necessity of pre-screening prior to image-text retrieval. As an example of the text-to-image retrieval task, the images with the dotted arrow are irrelevant to the textual query at the semantic level and are pre-screened out in advance.}
\label{fig1-1}
\end{figure}

In this paper, we present a novel keyword-guided pre-screening framework to solve the low efficiency problem of image-text retrieval.
We argue that a large number of gallery samples semantically irrelevant to the query can be screened out prior to the image-text retrieval algorithm, thus speeding up the following retrieval computation.
Taking the text-to-image retrieval as an example, as shown in Fig.~\ref{fig1-1}, given a textual query about the dog, it is unnecessary to compute the similarities with images of the boy or men, and we can pre-screen such image galleries.
Towards this end, we predict the keywords of the texts and images and compare them to enable a coarse-grained gallery pre-screening. 
To accomplish this, we transfer the keyword prediction into a multi-label classification problem and develop multi-label classifiers to predict the keywords. 
Correspondingly, two critical issues should be considered. 
% 强调keywords提取，对screening精度的影响，继而对结果的影响。
Firstly, since the obtained keywords can be viewed as a semantic expression of the sample on the discrete space and used for screening, the quality of keywords largely impacts the screening accuracy and the subsequent retrieval result.
Therefore, \textbf{the accuracy of keyword prediction is a critical issue}.
Secondly, the purpose of the pre-screening process is to improve the retrieval efficiency, yet the process itself can bring extra computational overhead, offsetting the efficiency to some extent. 
%A pre-screening process with a huge computational overhead is not applicable to improve retrieval efficiency. 
Therefore, \textbf{the efficiency of the pre-screening process is another critical issue}.
%\textbf{a lightweight and low-cost pre-screening process is another critical issue}.
Regarding to the first issue, we aim to improve the classifiers' quality by employing the booming image-text retrieval technique.
Specifically, we propose a multi-task learning scheme by appending the multi-label classifiers to the image-text retrieval network.
As a result, only two embedding layers for the multi-label classification are introduced in the proposed pre-screening framework.
Combined with the inverted index in the search engine~\cite{cambazoglu2016scalability}, the lightweight pre-screening is fulfilled only with $O(1)$ querying time complexity and hence the second issue is solved accordingly.
The proposed pre-screening framework can be applied prior to the common image-text retrieval methods to improve retrieval efficiency.
Beyond that, the keyword prediction, which is considered as a multi-label classification task, can be readily achieved by other state-of-the-art classification techniques, thereby further improving the proposed framework.
Fig.~\ref{fig1} (c) shows the simple flowchart of the proposed framework applied prior to the Late2Early method.
We achieve efficiency improvement ($\sim 4.1\times$ speed up) with little extra cost ($\sim + 1.8$ million modal parameters, $\sim + 2.5$e-6 minutes of online running), even while enhancing the accuracy performance ($+0.1\%$ Rank-1).

The main contributions of this paper are four folds.
(1) We focus on the relatively undervalued low efficiency problem of image-text retrieval and propose a simple yet effective keyword-guided pre-screening framework for improving retrieval efficiency correspondingly.
% introduce the inverted index concept in NLP into the image-text retrieval task for improving retrieval efficiency. A keyword-guided coarse-to-fine image-text retrieval framework is proposed accordingly. 
(2) We convert the keyword prediction into a multi-label classification task and further propose a multi-task learning scheme for lightweight and high-performance keyword prediction.
(3) We incorporate the inverted index in the search engine into the keyword matching for improving pre-screening efficiency.
(4) The proposed framework has robust compatibility. The experimental results on two public benchmarks show that it can be readily applied to almost all image-text retrieval methods to boost efficiency with only a minimal cost.

\section{Related Work}
%improving the efficiency and effectiveness of the cross-modal image-text retrieval with an eye to practice application.

%In general, the cross-modal retrieval architecture~\cite{wang2016learning} consists of three parts: image processing branch aiming to the visual feature extraction, text processing branch aiming to the textual feature extraction and fusion module aiming to the interaction between the visual feature and the texutal feature for computing similarity. According to the proportion of the fusion module in the architecture, existing cross-modal image-text retrieval methods can be roughly categorized into two lines: later-fused methods~\cite{wang2016learning,liu2017learning,wang2018learning,zhang2018deep} and early-fused methods.

We categorize the existing cross-modal image-text retrieval methods into three lines: late fusion, early fusion and efficiency-focused methods, and we brief them as follows.

\textbf{Late fusion methods} usually focus on developing various image and text processing techniques to extract feature representations of each modality separately, after which the two modalities interact with each other only with a loss function for training~\cite{zheng2020dual,sarafianos2019adversarial,park2020mhsan,huang2018learning,karpathy2014deep,radford2021learning}. 
For instance, Zheng \emph{et al.}~\cite{zheng2020dual} constructed an end-to-end dual-path convolutional neural network to learn the image and text global representations and proposed an instance loss for better weight initialization;
Park \emph{et al.}~\cite{park2020mhsan} explored the important components in the image and text independently by a multi-head self-attention network for local correspondence;
Radford \emph{et al.}~\cite{radford2021learning} recently proposed a contrastive dual-path pre-training network trained on a large-scale dataset of 400 million image-text pairs and achieved satisfactory performance on various downstream tasks, including the image-text retrieval task.
The late fusion methods yield the pairwise similarities by a simple inner product interaction during inference, promising retrieval efficiency. 
Nevertheless, due to the lack of information fusion across modalities when learning the feature representations, the performance of this kind of method is generally limited.

\textbf{Early fusion methods} pay more attention to information fusion between the visual and textual domains, and the image and text data are pre-processed separately~\cite{wang2019camp,lee2018stacked,liu2020graph,li2019visual,lee2019learning,zhang2021vinvl,li2020oscar,wang2020pfan++,zhang2022unified}. 
Researchers concentrated on the local correspondence across modalities in the early fusion methods for maximizing performance.
For example, the cross-attention mechanism~\cite{vaswani2017attention} is used for the message interaction between the regions of the image and the words of the text~\cite{wang2019camp,lee2018stacked};
the graph neural network~\cite{scarselli2008graph} is adopted to explore higher-order relational information between locals across modalities~\cite{liu2020graph,li2019visual,lee2019learning}.
In recent years, with the success of Transformer architecture~\cite{vaswani2017attention}, the large-scale pre-training methods~\cite{li2020oscar,zhang2021vinvl} have gradually become a central paradigm in vision-and-language (V+L) research and achieved many state-of-the-art results on V+L tasks, including the image-text retrieval task. 
The cross-modal information is usually fused early and freely by a single-stream transformer-like network in most pre-training methods, classified as the early fusion methods.
The information fusion across modalities benefits the semantic alignment between modalities, thus the early fusion method has an advantage on performance compared with the late fusion method. 
However, we need to quadratically compute the feature representation online in the early fusion method. It is, therefore, not practical to apply it in real-world scenarios.
%These methods [x] proposed various Transformer-like architecture. 
%trained on multiple large-scale V+L datasets for exlpoiting relationship between vision and language, and are essentially classified as the early-fused methods.
%The early-fused methods mine information of two modalities together rather than seperately and achieve state-of-the-art results compared to the later-fused methods,Due to the deep fusion of cross-modal information, the early-fused methods have been the leading paradigm for facilitating the performance development
%of the image-text retrieval task. However, there is a huge gap from laboratory experiments to real-world applications for the early-fused method due to its prohibitive reference cost. Particularly, as the current state-of-the-art paradigm, the cross-modal pretraining method is usually characterized by the bulky network architecture, aggravating the problem of high time complexity.
%The cross-modal image-text retrieval research is thriving in performance, 

\begin{figure*}
\begin{center}
   \includegraphics[width=1.0\linewidth, height=0.37\linewidth]{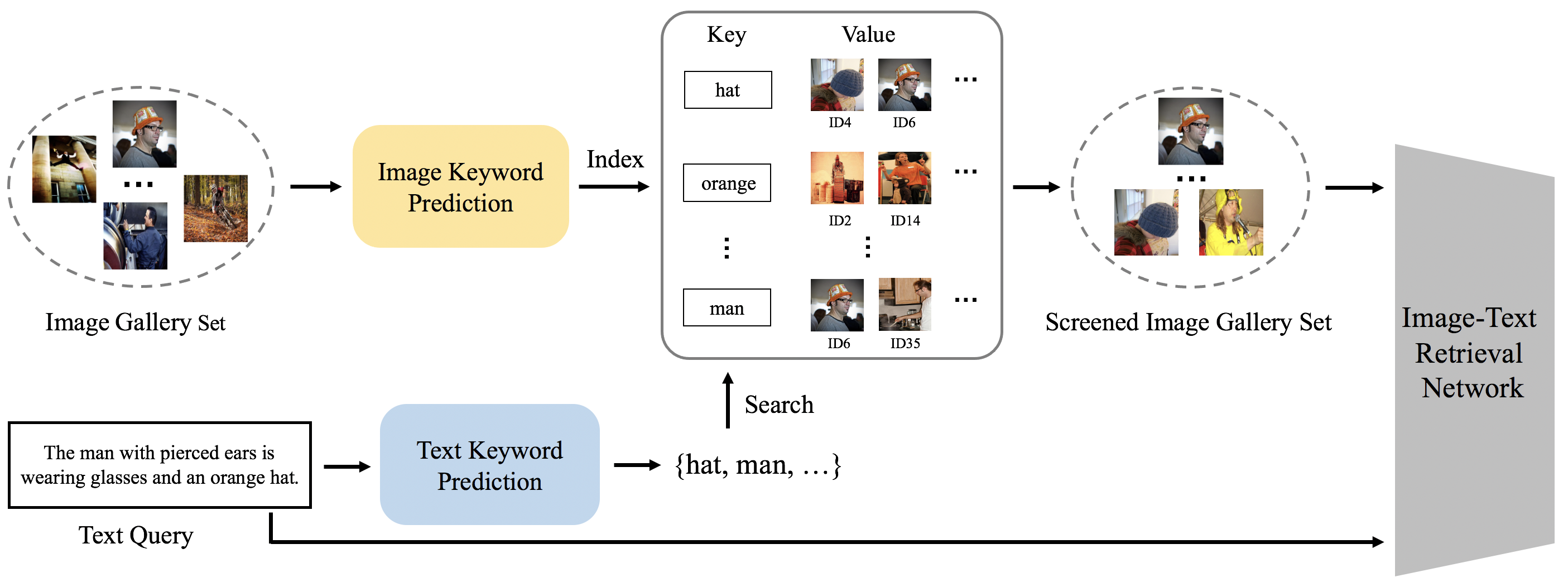}
\end{center}
   % \caption{A summary of the proposed framework; (b) and (c): An illustration of detail for the `Keyword Extraction' module in (a). The `Image-text Retrieval Network' module in (a) can be replaced with any of the early-fused methods or the Later2Early methods. For the image keyword extraction, we show the advanced image classification based on the image-text retrieval network ALBEF~\protect\cite{li2021align}.}
   \caption{A summary of the proposed framework with taking the text-to-image retrieval as an example. The `Image-Text Retrieval Network' module can be replaced with the common image-text retrieval method.}
\label{fig2}
\end{figure*}

%%%%%%%%% related work注释掉的正文
\textbf{Efficiency-focused methods} improve the image-text retrieval efficiency mainly from two perspectives. 
(1) Researchers optimized the model architecture smaller and lighter.
Yang \emph{et al.}~\cite{yang2017pairwise,yang2018shared,zhang2022watch} proposed to learn the hash codes of the feature representations with the help of the cross-modal hashing technique for saving memory and improving efficiency; 
Gan \emph{et al.}~\cite{gan2021playing} noticed that the over-parameterization issue results in the large memory and computation in the cross-modal pre-training models for the image-text retrieval, and proposed to search their sparse subnetwork with the aid of the lottery ticket hypothesis~\cite{frankle2018lottery}; 
similarly, Wang \emph{et al.}~\cite{wang2020minivlm} borrowed the MiniLM~\cite{wang2020minilm} structure to reduce the computation cost of the transformer module in the cross-modal pre-training models. 
And (2) some studies make efforts to yield a powerful late fusion model for narrowing down to a list of relevant galleries, within which the early fusion model is used to achieve the retrieval. 
For instance, Miech \emph{et al.}~\cite{miech2021thinking} distilled the knowledge of the early fusion model into the late fusion model, and adopted the distilled late fusion model for the preliminary screening of the gallery samples; 
Geigle \emph{et al.}~\cite{geigle2021retrieve} proposed to jointly train the early fusion model and the late fusion model with shared parameters for obtaining the high-quality late fusion model; 
Li \emph{et al.}~\cite{li2021align} introduced a contrastive loss on the feature representations from the image and text processing branches, and the similarity between the representations is used to select the relevant galleries for the following interaction module.
Unfortunately, the first group even with lightweight architecture still needs a long reference time due to the quadratic executions, and the second group has to perform an online search operation with $O(N)$ time complexity. 
Towards this end, there is further room for improving retrieval efficiency. 
Unlike the above two groups, we proposed a general low-cost pre-screening framework, which can be employed prior to them to further accelerate the retrieval.
%%%%%%%%% related work注释掉的正文

%\cite{lu2021visualsparta} aimed at solving text-to-image retrieval and proposed to purely and simply adopt independent word embedding of each token to represent the query, as a result of which the interaction matching across modalities can be pre-computed offline and the high-efficient retrieval is achieved at the expense of performance.

%Different from the first group for pursuing a smaller and lighter model, we aim for the optimization of the retrieval strategy by proposing a general pre- screening method, which can be applied in the first group for a super-efficient retrieval.
%Different from the second group for struggling a powerful later-fused model to screen out galleries, we develop a low-cost keyword-guided pre-screening method that can achieve the retrieval for big data in the reality effectively. 
%Furthermore, as a general first-stage screening framework, the proposed method can still be employed before the second group.

\section{Methodology}

For an image-text retrieval system, given a query sample in one modality, we need to retrieve a rank list from $N$ gallery samples in another based on the similarity with the query sample. 
% The state-of-the-art early-fused methods [xxx] compute the similarity by passing $N$ image-text combination through the network and come at prohibitive computational costs. The Later2Early methods [xxx] are developed subsequently by applying the later-fused network for the gallery screening before the early-fused network, which is still time-consuming with $O(N)$ time complexity.
The current state-of-the-art retrieval methods suffer from $N$-related complexity and come at prohibitive computational costs in reality.
Regarding this problem, we propose a keyword-guided pre-screening framework with minor computational overhead to narrow down to a list of $N_r$ relevant galleries ($N_r \ll N$) prior to the common image-text retrieval methods.
The proposed framework involves two modules: keyword prediction and pre-screening. 
At first, 
%the keywords of image and text, as a kind of semantic expression on the discrete space, are used for the gallery screening in the first coarse-grained stage. Thus the keyword's quality has important influence on the screening accuracy.
%the keywords in textual modality can be readily obtained by extracting the words in the text, yet the keywords in visual modality can not be obtained directly and need to an information transformation from the vision to the language.
%we extract the keyword in textual modality by NLTK~\cite{perkins2010python} and introduce the image classification into the keyword extraction in visual modality, referring to Section~\ref{ICKE}.
we present the multi-label classification technique for keyword prediction in visual and textual modalities, respectively, referring to Section~\ref{ICKE}.
%in which a vanilla image classification and an advanced image classification are developed separately
Thereafter, we introduce a keyword matching strategy based on the inverted index for the gallery screening in Section~\ref{KRP}.
%Finally, we proceed a whole image-text retrieval process in Section~\ref{TI}.
%screening out large amounts of easy-false galleries in Section.
%The complexity of the proposed method and the comparison with the Later2Early methods are analyzed theoretically in Section~\ref{CAD}.

The block diagram of the proposed framework is illustrated in Fig.~\ref{fig2}.
The proposed framework can be embedded into almost all common image-text retrieval networks.
Combining the proposed coarse-grained pre-screening and the state-of-the-art fine-grained retrieval brings a superior tradeoff between accuracy and efficiency.

%-------------------------------------------------------------------------

\subsection{Keyword Prediction}
\label{ICKE}
%We achieve the gallery screening by matching the keywords across modalities, thus the keywords of image and text are firstly need to be extracted.
%The keywords in textual modality are readily obtained by extracting the words in the text, yet the ones in visual modality can not be obtained directly.

\textbf{Visual Modality.}
We train a multi-label image classifier for predicting the image's keywords. 
Any common multi-label classification techniques~\cite{ridnik2021asymmetric,lanchantin2021general} can be adopted.
However, we can not directly adopt the classifier trained on the multi-label classification benchmarks, such as COCO-80 and VG-500~\cite{lanchantin2021general}. 
In doing so, the image keywords obtained from the class annotations in these benchmarks may be misaligned with the actual semantic content in the image due to the data gap between the image-text retrieval and the multi-label classification, thereby causing the imprecise keyword-guided screening. 
To this end, we train the classifier based on the image-text retrieval benchmarks.
Specifically, the image's ground-truth annotations are generated from the paired texts.
Referring to the setting in the classification task where the class annotations are generally nouns, we use the nouns in the paired texts as the image annotations.
Alternatively, we also experimentally use the nouns, verbs and adjectives from the paired texts as the ground-truth annotations. 
Better performance results are obtained by only using the nouns as the annotations.

\begin{figure}
\begin{center}
   \includegraphics[width=1.0\linewidth, height=0.65\linewidth]{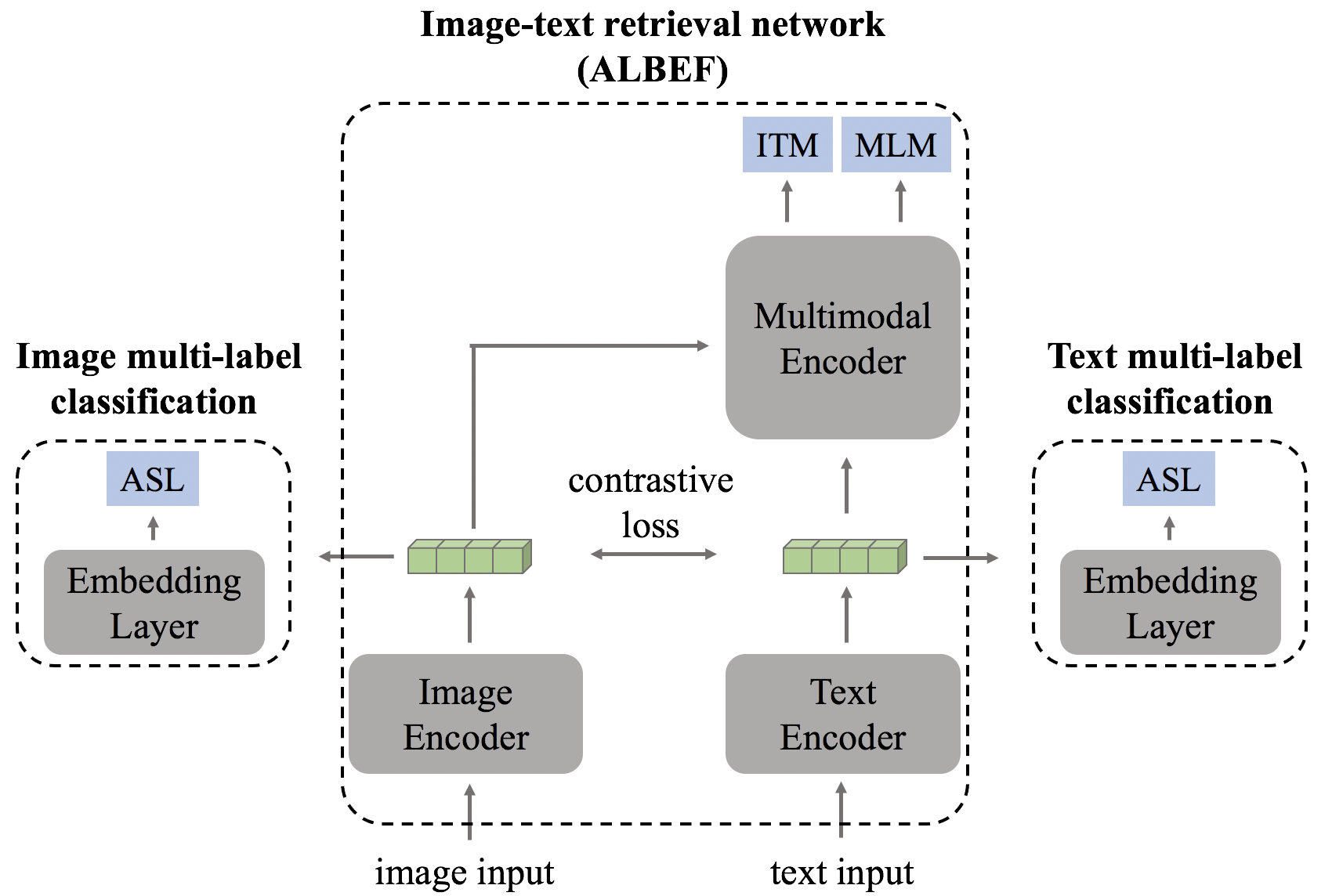}
\end{center}
   % \caption{A summary of the proposed framework; (b) and (c): An illustration of detail for the `Keyword Extraction' module in (a). The `Image-text Retrieval Network' module in (a) can be replaced with any of the early-fused methods or the Later2Early methods. For the image keyword extraction, we show the advanced image classification based on the image-text retrieval network ALBEF~\protect\cite{li2021align}.}
   \caption{Illustration of the proposed advanced classifiers appended in the ALBEF~\protect\cite{li2021align} for keyword prediction.}
\label{fig3}
\end{figure}

Beyond adopting the existing common classification technique for image keyword prediction, we further propose an advanced classification to enhance the classifier’s performance and the follow-up screening accuracy and retrieval result.   
%influences screening accuracy, ultimately affecting the image-text retrieval result.
%for more accurate gallery screening. 
The multi-label image classification task can essentially be viewed as an image-to-label retrieval task, sharing certain characteristics with the image-text retrieval task. 
We believe that the booming image-text retrieval technique can positively contribute to the multi-label image classification on performance. 
Thereby, we propose a multi-task learning scheme by appending the multi-label classifier to the image-text retrieval network. 
Specifically, after the image processing branch in the retrieval network, we add an extra label embedding layer with a multi-label classification loss, thus enabling the multi-task learning of the image-text retrieval and the multi-label classification.
As an accessory, the advanced classification introduces minimal computational overhead into the proposed pre-screening framework compared to the common classification.

For the classification loss, rather than adopting a frequently-used binary cross-entropy loss~\cite{lanchantin2021general}, we adopt a state-of-the-art asymmetric loss (ASL)~\cite{ridnik2021asymmetric}.
Compared to the binary cross-entropy loss, the ASL loss operates dynamically on positive and negative samples during training and considers the positive-negative imbalance problem in the classification task.
Specifically, 
\begin{equation}\label{e1}
L= \sum_{i=1}^{l}-y_{i}L_{+}-\left ( 1-y_{i} \right )L_{-},
\end{equation}
where $y_{i}=1$ represents that the $i$-th class annotation is the ground-truth of the image $x$ and vice versa, and  
\begin{equation}\label{e2}
\begin{cases}
L_{+}= \left ( 1-p_{i} \right )^{\alpha _{+}}\log\left ( p_{i} \right ) \\
L_{-}= \left ( \tilde{p}_i \right )^{\alpha _{-}}\log\left ( 1-\tilde{p}_i \right ),
\end{cases}
\end{equation}
where $\alpha _{+}$ and $\alpha _{-}$ are the positive and negative focusing parameters, respectively.
With a high $\alpha _{+}$ (resp. $\alpha _{-}$), the contribution from easy positives with $p_i \gg 0.5$ (resp. easy negatives with $p_i \ll 0.5$) to the loss is weakened, leading to more focus on more challenging samples in training. 
$\tilde{p}_i=max\left ( p{_i}- \Delta, 0 \right )$ is a shifted label probability, and the negative sample will be discarded when $\tilde{p}_i<\Delta$.

\begin{figure}
\begin{center}
   \includegraphics[width=1\linewidth, height=0.72\linewidth]{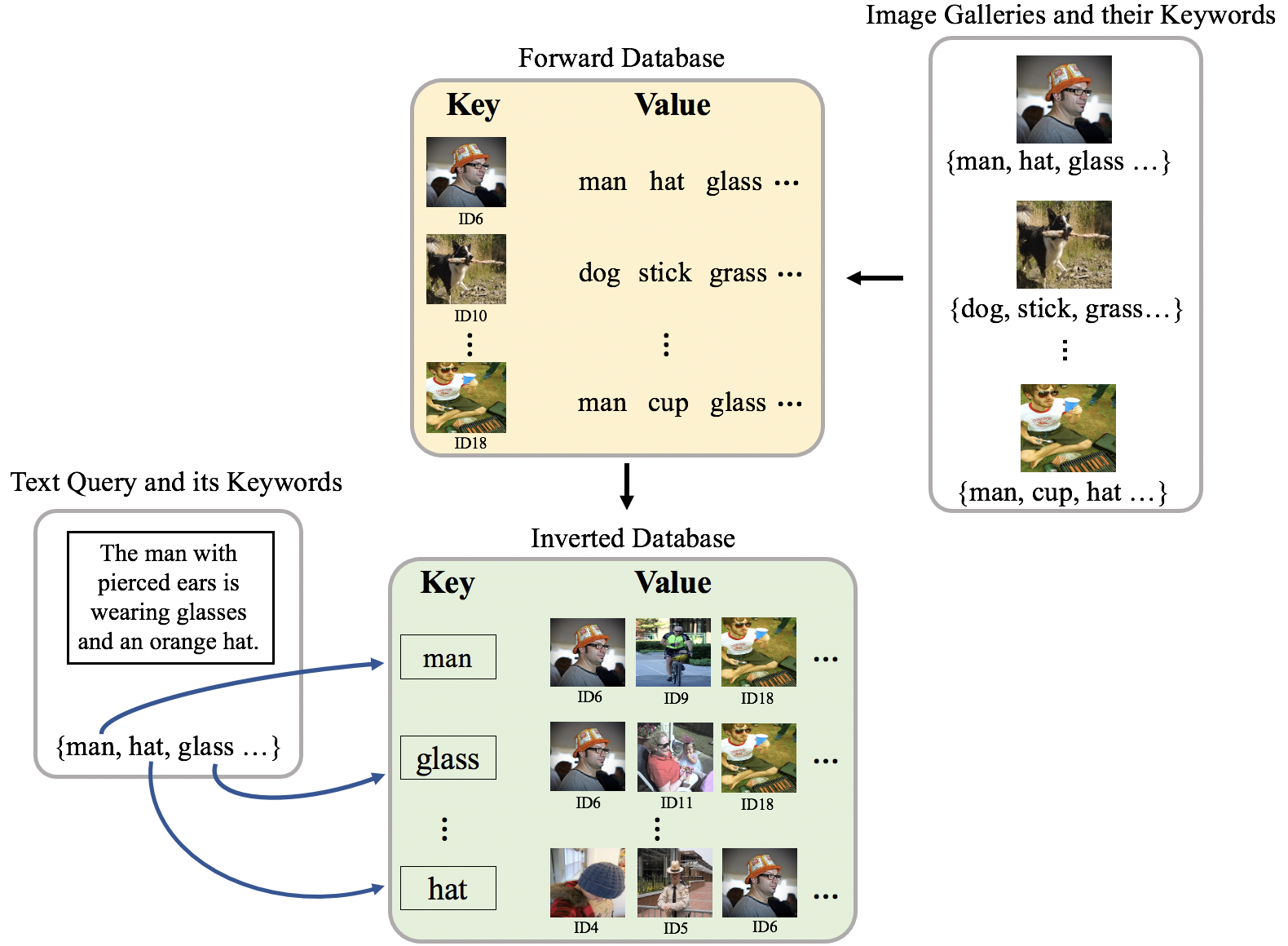}
\end{center}
   % \caption{A summary of the proposed framework; (b) and (c): An illustration of detail for the `Keyword Extraction' module in (a). The `Image-text Retrieval Network' module in (a) can be replaced with any of the early-fused methods or the Later2Early methods. For the image keyword extraction, we show the advanced image classification based on the image-text retrieval network ALBEF~\protect\cite{li2021align}.}
   \caption{Illustration of the pre-screening with the inverted index. We take text-to-image retrieval as an example.}
\label{fig4}
\end{figure}

\textbf{Textual Modality.}
Intuitively, we can extract the words directly from the text by using the natural language toolkit~\cite{perkins2010python} as its keywords.
As a result, the inference text's keywords are derived from the inference data, and yet the image's keywords are still from the class annotations of the training data.
The gap between the training and inference data tends to bring a possible situation with no overlap between the keywords of text and image in inference, thus leading to the failure of pre-screening.
Alternatively, we can introduce a multi-label classification into the keyword prediction in textual modality, as in viual one.
The text's ground-truth annotations for training are equivalent to the ones of the paired images, ensuring the overlap of keywords across modalities in inference and avoiding the pre-screening's failure.
%The text's ground-truth annotations for training are obtained by extracting the nouns in this text and its positive textual samples, which are equivalent to the ones of the paired images, ensuring the overlap of keywords across modalities in inference and avoiding the pre-screening's failure.
Yet at the same time, the classifier is likely to predict the wrong keywords for the text.
From the above analysis, merging the predicted labels and extracted nouns to the keywords is a complementary solution.
We experimentally study the performance of the above models for generating text keywords.
The multi-label classification results in better retrieval performance and is used in the proposed framework.
Beyond using the common classification technique, we append a multi-label classifier after the text processing branch in the retrieval network for the advanced classifier.

We take an image-text retrieval method ALBEF~\cite{li2021align} as an example for the two advanced classifiers, as illustrated in Fig.~\ref{fig3}.
The label embedding layer together with the ASL are appended to the end of the image encoder and text one, respectively.
In inference, the class annotations with top-$R_I$ and top-$R_T$ highest probabilities are used as the image and text keywords, respectively.

%It is worth mentioning that the inference text's keywords are derived from the inference data, yet the images' keywords are still from the class annotations of the training data. Due to the minor gap between the training and inference data, there may be no overlap between the text's keywords and the image's keywords. It would cause the failure of pre-screening for such texts, for which we skip over the pre-screening stage. There are only $37$ and $9$ such texts in Flickr30K and MS-COCO, respectively.
%For every image, we select the annotations with $K=20$ highest probabilities as its keywords.

\subsection{Keyword Matching for Pre-screening}
\label{KRP}
% \subsubsection{Coarse-grained Screening}
%We focus on the efficiency improvement by pre-screening out the massive irrelevant galleries before the standard image-text retrieval methods, which can also be called a coarse-grained pre-screening stage.
%The sample's keywords abstract its information and can be interpreted as its entries, used as guidance for gallery pre-screening.
The sample information is abstracted as the keywords expressed in discrete form and used as guidance for gallery pre-screening prior to the image-text retrieval network. 
Intuitively, we can compare the keywords of the query and each gallery sample and screen out the galleries without overlap with the query regarding the keywords.
However, it has a high computational complexity in reality due to the traversal operator in the gallery set with a tremendous number of $N$.
For this, we achieve the high-efficient pre-screening with the inverted index technique, inspired by the search engine~\cite{cambazoglu2016scalability}.
In particular, the pre-screening involves an index step that builds the mapping between the keywords and the gallery samples, a search step that picks out the gallery samples sharing the same keywords with the query sample and discards the rest of the gallery samples.

\textbf{Index.} 
%Many search engines incorporate an inverted index when evaluating a search query to quickly locate documents containing the words in a query and then rank these documents by relevance. Because the inverted index stores a list of the documents containing each word, the search engine can use direct access to find the documents associated with each word in the query in order to retrieve the matching documents quickly. The following is a simplified illustration of an inverted index: 
After the keyword prediction, we have obtained a mapping from the gallery to the keyword.
It implies that a naive forward database index has been implicitly built where the gallery sample ID is specified as the key and its keywords as the value, as shown in Fig.~\ref{fig4}. 
%Given the query's keywords, we need to step through each key and compare with the associated value, leading to a time-consuming pre-screening.
A time-consuming pre-screening is required based on the database.
Consequently, we construct an inverted database index in which the keyword is specified as the key and its paired gallery samples as the value, resulting in a fast pre-screening as follows.

\textbf{Search.}
As shown in Fig.~\ref{fig4}, with the inverted index created, the pre-screening can be resolved by two steps:
a query step with quickly jumping to the keys having the same as the query's keywords, and a mergence step with merging the associated values to as the retaining gallery samples.

\section{Complexity Analysis and Discussion}
\label{CAD}
While achieving efficiency improvement, the proposed framework itself can bring extra resource overhead. We analyze its time and space complexities and compare it with the Late2Early method, which is closely related to the proposed framework.

%\textbf{Complexity on Coarse-grained Stage.}
%\textbf{Complexity Analysis.}
%For the proposed framework, we can first compute the keywords of all galleries and build the inverted index offline. The space complexity of storing the inverted index is about $O(Nl)$, where $l$ is the mean number of the keywords for each gallery. In general, $l\ll 100$ and is in single or double digits. Then when inputting a query in real-time, its keywords are computed by the syntactic parsing (text as query) or the classifier (image as query) and we launch the search step in the coarse-grained screening stage, in which the time complexity depends on the index merging operation and outside matters are $O(1)$ complexity. The time complexity of the vanilla index merging operation is $O(k)$, where $k \ll N$ is the number of screened galleries by the coarse-grained stage. Furthermore, the index merging operation can be quickly implemented by means of the database technology with the time complexity being less than $O(k)$
The online computation is composed of query processing and gallery screening. 
In the proposed framework, there is no extra time to spend on the query processing (i.e., query keyword prediction) due to the proposed multi-task learning scheme;
during the gallery screening, with an $O(1)$ querying time complexity, the next merging step with far less than $O(N)$ complexity dominates the time complexity of the screening.
In the Late2Early method, given a real-time query, its feature is extracted with high complexity and $k$-nearest neighbors from the gallery samples are selected with $O(N)$ time complexity.

For the space complexity, we only need to add two label embedding layers and store the inverted index only including ID numbers, giving a clear advantage over the Late2Early method, usually with two heavy visual and textual encoding networks and storing the gallery samples' features for the gallery screening.   

In addition, the proposed framework can be applied prior to almost all image-text retrieval methods to improve retrieval efficiency.
When applied prior to the Late2Early method, the proposed pre-screening selects $N_r$ gallery samples ($N_r\ll N$) to the follow-up Late2Early method, bringing $O(N_r)$ time complexity instead of the original $O(N)$ in the Late2Early method.
When applied prior to the other method, the proposed pre-screening plays a similar role as the late fusion network in the Late2Early method, yet offers a significant advantage over the resource overhead compared with the Late2Early methods.

\section{Experiments}

\subsection{Experiment Settings}

%The digit following `@' in column `Method' represents the number of the galleries to the early-fused network in the Later2Early method or the parameter $R$ for extracting image keywords in the proposed framework.

\textbf{Datasets.}
We conduct experiments on two widely-used image-text retrieval benchmark datasets: Flickr30K~\cite{plummer2015flickr30k} and MS-COCO~\cite{lin2014microsoft}, containing $31,014$ and $123,287$ images, respectively. Both datasets have five associated textual descriptions per image. 
Following the common settings~\cite{chen2021learning,kim2021vilt}, we split Flickr30K into $29,000$ images for training, $1,014$ for validation and $1,000$ for inference; we use $113,287$ images for training, $5,000$ for validation and $5,000$ for inference in MS-COCO. 
We report the results for both image-to-text retrieval (TR) and text-to-image retrieval (IR) in the experiments.

\begin{table*}[tb!]
\caption{The results of the proposed framework applied prior to various image-text retrieval methods. The symbols `+' and `++' in the Time(m) and Para(M) columns refer to the online running time and the number of modal parameters resulted from the method in the corresponding row. The symbol `$\times$' in the Speedup column refers to the retrieval acceleration of the method in the previous row and the `$\times\times$' refers to the screening acceleration of the method in the previous row. For clarity, we present the R@(sum) variation of the proposed framework compared to the method in the top row above `+Ours' or `++Ours'.}
         \begin{center}
         \renewcommand\arraystretch{1.8}
\resizebox{1\textwidth}{!}{
         \begin{tabular}{l|l||c c c c c c | c c | c c c c c c| c c | c}
         \Xhline{1.3pt}
         {\multirow{3}{*}{Genre}} & {\multirow{3}{*}{Method}} & \multicolumn{8}{c|}{Flickr30K} & \multicolumn{8}{c|}{MS-COCO}  &  \\
         \cline{3-19}
         &  & \multicolumn{3}{c}{TR} & \multicolumn{3}{c|}{IR} & {\multirow{2}{*}{Time(m)}} & {\multirow{2}{*}{Speedup}} & \multicolumn{3}{c}{TR} & \multicolumn{3}{c|}{IR} & {\multirow{2}{*}{Time(m)}} & {\multirow{2}{*}{Speedup}} & {\multirow{2}{*}{Para(M)}} \\
         &  & R@1 & R@5 & R@sum & R@1 & R@5 & R@sum &   &   & R@1 & R@5 & R@sum & R@1 & R@5 & R@sum &   &   &  \\ 
          \hline
         Early fusion & ALBEF$_{all}$ & 95.8 & 99.7 & 195.5 & 84.3 & 95.9 & 180.2 & 2.2e+3  & ~~- & 77.6 & 94.3 & 171.9 & 60.6 & 84.2 & 144.8 & 5.5e+4 & ~~- & 56.7 \\
         Late2Early & +ALBEF$_{0}$ & 95.9 & 99.8 & 195.7 & 85.5 & 97.5 & 183.0 & +1.6e+1 & 23.4$\times$ & 77.6 & 94.3 & 171.9 & 60.7 & 84.3 & 145.0 & +7.8e+1 & 117.2$\times$  & +152.9 \\ 
         & ++Ours & 95.9 & 99.8 & 195.7 (\textcolor[rgb]{0,1,0}{$\rightarrow$}) & 85.2 & 97.0 & 182.2 (\textcolor[rgb]{1,0,0}{$\downarrow0.8$}) & ++4.2e-3 & 2.1$\times\times$ & 77.6 & 94.1 & 171.7 (\textcolor[rgb]{1,0,0}{$\downarrow0.2$}) & 60.5 & 84.0 & 144.5 (\textcolor[rgb]{1,0,0}{$\downarrow0.5$}) & ++6.2e-2 & 4.1$\times\times$ & ++1.3 \\
         \hline
         Early fusion & ViLT-B/32a & 83.7 & 96.8 & 180.5 & 64.3 & 88.8 & 153.1 & 5.9e+3 & ~~- & 61.6 & 86.4 & 148.0 & 42.8 & 72.9 & 115.7 & 1.5e+5 & ~~- & 111.6 \\
         Late2Early & +ALBEF$_{0}$ & 85.6 & 97.2 & 182.8 & 66.7 & 90.1 & 156.8 & +1.6e+1 & 23.4$\times$ & 65.6 & 89.5 & 155.1 & 46.0 & 75.8 & 121.8 & +7.8e+1 & 117.2$\times$ & +152.9 \\
         & ++Ours & 86.4 & 98.4 & 184.8 (\textcolor[rgb]{0,1,0}{$\uparrow2.0$}) & 66.2 & 90.2 & 156.4 (\textcolor[rgb]{1,0,0}{$\downarrow0.4$}) & ++4.2e-3 & 2.1$\times\times$ & 65.6 & 89.4 & 155.0 (\textcolor[rgb]{1,0,0}{$\downarrow0.1$}) & 46.1 & 75.3 & 121.4 (\textcolor[rgb]{1,0,0}{$\downarrow0.4$}) & ++6.2e-2 & 4.1$\times\times$ & ++1.3 \\
         \hline
         Early fusion & ViLT-B/32a & 83.7 & 96.8 & 180.5 & 64.3 & 88.8 & 153.1 & 5.9e+3 & ~~- & 61.6 & 86.4 & 148.0 & 42.8 & 72.9 & 115.7 & 1.5e+5 & ~~- & 111.6 \\
         Late2Early & +LightDOT & 85.5 & 97.8 & 183.3 & 66.2 & 90.2 & 156.4 & +1.4 & 23.4$\times$ & 65.3 & 88.7 & 154.0 & 45.9 & 75.6 & 121.5 & +6.9 & 117.2$\times$ & +222.9 \\
         & ++Ours & 85.7 & 97.9 & 183.6 (\textcolor[rgb]{0,1,0}{$\uparrow0.3$}) & 65.6 & 90.5 & 156.1 (\textcolor[rgb]{1,0,0}{$\downarrow0.3$}) & ++4.2e-3 & 2.1$\times\times$ & 65.3 & 88.5 & 153.8 (\textcolor[rgb]{1,0,0}{$\downarrow0.2$}) & 45.6 & 74.8 & 120.4 (\textcolor[rgb]{1,0,0}{$\downarrow1.1$}) & ++6.2e-2 & 4.1$\times\times$ & ++1.3 \\
         \hline
         Late fusion & LightDOT & 83.9 & 97.2 & 181.1 & 69.9 & 91.1 & 161.0 & 1.4 & ~~- & 60.0 & 85.2 & 145.2 & 45.8 & 74.5 & 120.3 & 6.9 & ~~- & 222.9 \\
         & +Ours & 84.9 & 97.9 & 182.8 (\textcolor[rgb]{0,1,0}{$\uparrow1.7$}) & 69.3 & 90.6 & 159.9 (\textcolor[rgb]{1,0,0}{$\downarrow1.1$}) & +4.2e-3 & 2.1$\times$ & 60.4 & 85.3 & 145.7 (\textcolor[rgb]{0,1,0}{$\uparrow0.5$}) & 45.9 & 74.7 & 120.6 (\textcolor[rgb]{0,1,0}{$\uparrow0.3$}) & +6.2e-2 & 4.1$\times$ & +1.3 \\ 
         \hline
         Late fusion & CLIP & 84.9 & 97.2 & 182.1 & 64.8 & 87.2 & 152.0 & 2.7 & ~~- & 56.1 & 79.0 & 135.1 & 36.1 & 60.9 & 97.0 & 13.2 & ~~- & 427.6 \\
         & +Ours & 84.9 & 97.3 & 182.2 (\textcolor[rgb]{0,1,0}{$\uparrow0.1$}) & 65.0 & 87.2 & 152.2 (\textcolor[rgb]{0,1,0}{$\uparrow0.2$}) & +4.2e-3 & 2.1$\times$ & 56.1 & 79.0 & 135.1 (\textcolor[rgb]{0,1,0}{$\rightarrow$}) & 36.2 & 61.1 & 97.3 (\textcolor[rgb]{0,1,0}{$\uparrow0.3$}) & +6.2e-2 & 4.1$\times$ & +1.3 \\ 
         \hline
         Early fusion & ViLT-B/32a & 83.7 & 96.8 & 180.5 & 64.3 & 88.8 & 153.1 & 5.9e+3 & ~~- & 61.6 & 86.4 & 148.0 & 42.8 & 72.9 & 115.7 & 1.5e+5 &  ~~- & 111.6 \\
         & +Ours & 83.7 & 96.8 & 180.5 (\textcolor[rgb]{0,1,0}{$\rightarrow$}) & 65.2 & 89.0 & 154.2 (\textcolor[rgb]{0,1,0}{$\uparrow1.1$}) & +4.2e-3 & 2.1$\times$ & 61.9 & 86.5 & 148.4 (\textcolor[rgb]{0,1,0}{$\uparrow0.4$}) & 44.8 & 73.7 & 118.5 (\textcolor[rgb]{0,1,0}{$\uparrow2.8$}) & +6.2e-2 & 4.1$\times$ & +1.3 \\
          \Xhline{1.2pt}
         \end{tabular}
         }  
         \label{tab1}
         \end{center}
\end{table*}

\textbf{Evaluation metrics.}
We adopt the widely used metric Rank-k (R@k) for evaluation.
R@k represents the expectation of correct match at the $k$-th in the ranking list and we report R@1, R@5 and R@sum (\emph{i.e.}, the sum of R@1 and R@5).
In addition, in order to well reveal the method's effectiveness, we report the number of modal parameters (Para. for short, in millions), the online running time\footnote{The running time is measured without acceleration operation for a fair comparison.} (in minutes) of the modal for all queries on a dataset and the speedup ratio for retrieval. 
For the running time and the speedup ratio, we report the mean average for TR and IR.
It is worth noting that the number of modal parameters in the proposed framework is different on different datasets, since there are only two label embedding layers in the proposed framework and its modal parameter depends on the number of labels\footnote{There are $0.8$ and $1.8$ million parameters for the proposed framework on Flickr30K and MS-COCO, respectively.}. We report the average of the values on Flickr30K and MS-COCO datasets. 

% the fps describes how many results of the query samples can be output in a second. 
%and is reported based on the CPU due the unstable running time on GPU.
 
\textbf{Implementation details.}
%提一下，由于训练标签和测试标签的gap，导致一些样本无法参与pre-screening，我们目前处理方式是直接进入下一阶段。
The experiments are conducted on eight GeForce RTX 3090 GPUs with 24 GB of memory.
For the advanced classification, we append the classifiers to the image-text retrieval network ALBEF~\cite{li2021align} in the experiments.
Based on the trained ALBEF on Flickr30K (resp. MS-COCO), we continue to train the classifier for 10 (resp. 5) epochs with a batch size of 128.
For training the classifier better, we remove the class annotations paired with fewer than $100$ images.
%considering that some class annotations are paired with few images and are unfavorable for the classifier's training, we select the class annotations paired with greater than $100$ images as the final class annotations. 
As a result, there are $539$ and $1,122$ class annotations in Flickr30K and MS-COCO, respectively.
We set $\alpha_{+}=0$, $\alpha_{-}=3$ and $\Delta=0.05$ for ASL loss referring to the settings in \cite{ridnik2021asymmetric}, and set $R_I=15$ and $R_T=3$ in the image and text classifications, respectively.
%It is worth mentioning that the inference text's keywords are derived from the inference data, yet the images' keywords are still from the class annotations of the training data. Due to the minor gap between the training and inference data, there may be no overlap between the text's keywords and the image's keywords. It would cause the failure of pre-screening for such texts, for which we skip over the pre-screening stage. There are only $37$ and $9$ such texts in Flickr30K and MS-COCO, respectively.
%For every image, we select the annotations with $K=20$ highest probabilities as its keywords.

\textbf{Baselines.}
The proposed framework can be applied prior to the common image-text retrieval method, \emph{e.g.}, the early fusion, late fusion or the Late2Early methods.
Several image-text retrieval methods are involved as the baselines in the experiments:
the early fusion methods ALBEF$_{all}$ and ViLT-B/32a~\cite{kim2021vilt}; 
the late fusion methods ALBEF$_{0}$, LightDOT~\cite{sun2021lightningdot} and CLIP~\cite{radford2021learning};
the Late2Early methods from the free combination of the above late fusion and early fusion baselines\footnote{Due to unpublished codes, the pure Late2Early methods~\cite{miech2021thinking,sun2021lightningdot,liu2021inflate} can not be used in the experiments.}.
Referring to the setting in \cite{li2021align}, we select $128$ gallery samples by the late fusion network and fed them into the early fusion network in the Late2Early.
Specifically, ALBEF$_{all}$ and ALBEF$_{0}$ are evolved from the ALBEF~\cite{li2021align}.
In the ALBEF, given one query, all gallery samples first get through the image/text processing branch to compute the similarities with the query, and then the top-$128$ gallery samples together with the query are sent to the following interaction module for retrieval.
ALBEF$_{all}$ refers to all galleries instead of top-$k$ ones being sent to the interaction module and is essentially an early fusion method, and ALBEF$_{0}$ refers to that there are no galleries to the interaction module and is essentially a late fusion method.
We report the baselines' results by running the published code in the experiments.

\subsection{Results and Discussion}

We report the results of the proposed pre-screening framework applied prior to the Late2Early, late fusion or early fusion methods in Table~\ref{tab1}.
%We show the fps and Para for the gallery screening, and the fps is reported based on the CPU due to the unstable running time on GPU.
%The number of the retaining galleries to the early-fused network in the Later2Early method and the parameter $R$ for extracting image keywords in the proposed framework can impact the results, thus we report the results with different parameters where the digit following `@' represents the parameter. More results of the proposed framework with various different $R$ are reported in Section~\ref{FA}.
No matter what type of method is applied, the proposed framework can achieve acceleration, while keeping the performance and sometimes offering a remarkable improvement.
For example, a $2.0\%$ increase at R@sum is achieved by the proposed framework applied prior to the ALBEF$_{0}$+ViLT-B/32a method on Flickr30K (TR), and a $2.8\%$ increase is achieved applied prior to the ViLT-B/32a method on MS-COCO (IR).
The price for this is only the $4.2$e-3 (resp. $6.2$e-2) minutes of online running time on Flickr30K (resp. MS-COCO) and an average of $1.3$ million modal parameters.
It is worth noting that the proposed pre-screening framework working on the early fusion method plays the same role as the late fusion network in the Late2Early method, that is, both aim at improving the retrieval efficiency of the early fusion method.
Targeting the acceleration of the same early fusion method (\emph{i.e.}, ViLT-B/32a), the cost of the proposed framework is far less than that of the late fusion network (\emph{i.e.}, ALBEF$_{0}$ and LightDOT). 
Specifically, the proposed framework is an average of $2,072$ (resp. $685$) times faster than the late fusion network on Flickr30K (resp. MS-COCO) for the running time and $145$ times less than the late fusion network for the modal parameter.

\begin{table*}[tb!]
\caption{Comparison with the ANN-based Late2Early methods. We use the ANN algorithms: product quantization (PQ) and PQ with inverted file (IVFPQ). The symbol `$\times$' in the Speedup column refers to the screening acceleration of the Late2Early method. For clarity, we present the R@(sum) variation of the method compared to the Late2Early method.}
         \begin{center}
         \renewcommand\arraystretch{1.8}
\resizebox{1\textwidth}{!}{
         \begin{tabular}{l|l||c c c c c c c| c c c c c c c}
         \Xhline{1.3pt}
         {\multirow{3}{*}{Genre}} & {\multirow{3}{*}{Method}} & \multicolumn{7}{c|}{Flickr30K} & \multicolumn{7}{c}{MS-COCO} \\
         \cline{3-16}
         & & \multicolumn{3}{c}{TR} & \multicolumn{3}{c}{IR} & {\multirow{2}{*}{Speedup}} & \multicolumn{3}{c}{TR} & \multicolumn{3}{c}{IR} & {\multirow{2}{*}{Speedup}} \\
         & & R@1 & R@5 & R@sum & R@1 & R@5 & R@sum &  & R@1 & R@5 & R@sum & R@1 & R@5 & R@sum & \\
         \hline
         Late2Early & ALBEF$_{0}$+ALBEF$_{all}$ & 95.9 & 99.8 & 195.7 & 85.5 & 97.5 & 183.0 & ~~- & 77.6 & 94.3 & 171.9 & 60.7 & 84.3 & 145.0 & ~~- \\
         ANN Late2Early & +PQ & 95.7 & 99.7 & 195.4 (\textcolor[rgb]{1,0,0}{$\downarrow0.3$}) & 66.0 & 72.5 & 138.5 (\textcolor[rgb]{1,0,0}{$\downarrow44.5$}) & 1.3$\times$ & 75.9 & 92.0 & 167.9 (\textcolor[rgb]{1,0,0}{$\downarrow4.0$}) & 26.2 & 30.0 & 56.2 (\textcolor[rgb]{1,0,0}{$\downarrow88.8$}) & 2.2$\times$ \\
         ANN Late2Early & +IVFPQ & 76.5 & 80.8 & 157.3 (\textcolor[rgb]{1,0,0}{$\downarrow38.4$}) & 57.2 & 62.2 & 119.4 (\textcolor[rgb]{1,0,0}{$\uparrow63.6$}) & 3.2$\times$ & 56.5 & 71.1 & 127.6 (\textcolor[rgb]{1,0,0}{$\downarrow44.3$}) & 45.3 & 62.4 & 107.7 (\textcolor[rgb]{1,0,0}{$\downarrow37.3$}) & 3.6$\times$ \\
         & +Ours & 95.9 & 99.8 & 195.7 (\textcolor[rgb]{0,1,0}{$\rightarrow$}) & 85.2 & 97.0 & 182.2 (\textcolor[rgb]{1,0,0}{$\downarrow0.8$}) & 2.1$\times$ & 77.6 & 94.1 & 171.7 (\textcolor[rgb]{1,0,0}{$\downarrow0.2$}) & 60.5 & 84.0 & 144.5 (\textcolor[rgb]{1,0,0}{$\downarrow0.5$}) & 4.1$\times$ \\
         \hline
         Late2Early & ALBEF$_{0}$+ViLT-B/32a & 85.6 & 97.2 & 182.8 & 66.7 & 90.1 & 156.8 & ~~- & 65.6 & 89.5 & 155.1 & 46.0 & 75.8 & 121.8 & ~~- \\
         ANN Late2Early & +PQ & 86.0 & 97.9 & 183.9 (\textcolor[rgb]{0,1,0}{$\uparrow1.1$}) & 55.9 & 70.6 & 126.5 (\textcolor[rgb]{1,0,0}{$\downarrow30.3$}) & 1.3$\times$ & 64.9 & 87.5 & 152.4 (\textcolor[rgb]{1,0,0}{$\downarrow2.7$}) & 22.8 & 29.4 & 52.2 (\textcolor[rgb]{1,0,0}{$\downarrow69.6$}) & 2.2$\times$ \\
         ANN Late2Early & +IVFPQ & 74.0 & 83.8 & 157.8 (\textcolor[rgb]{1,0,0}{$\downarrow25.0$}) & 50.2 & 61.4 & 111.6 (\textcolor[rgb]{1,0,0}{$\downarrow45.2$}) & 3.2$\times$ & 53.5 & 74.0 & 127.5 (\textcolor[rgb]{1,0,0}{$\downarrow27.6$}) & 35.4 & 56.6 & 92.0 (\textcolor[rgb]{1,0,0}{$\downarrow29.8$}) & 3.6$\times$ \\
         & +Ours & 86.4 & 98.4 & 184.8 (\textcolor[rgb]{0,1,0}{$\uparrow2.0$}) & 66.2 & 90.2 & 156.4 (\textcolor[rgb]{1,0,0}{$\downarrow0.4$}) & 2.1$\times$ & 65.6 & 89.4 & 155.0 (\textcolor[rgb]{1,0,0}{$\downarrow0.1$}) & 46.1 & 75.3 & 121.4 (\textcolor[rgb]{1,0,0}{$\downarrow0.4$}) & 4.1$\times$ \\
          \Xhline{1.2pt}
         \end{tabular}}
         \label{tab2}
         \end{center}
\end{table*}

\subsection{Further Analysis}
\label{FA}
\textbf{Comparison with ANN-based Late2Early method.}
Though the Late2Early method suffers from an $O(N)$ screening time complexity, the Approximate Nearest Neighbor search (ANN)~\cite{arya1998optimal} can be applied to speed it up.
Two commonly used ANN algorithms from the Facebook AI Similarity Search (FAISS) library, \emph{i.e.}, the Product Quantization (PQ) and the Product Quantization with InVerted File (IVFPQ), are applied in the Late2Early for comparing with the proposed framework.
The comparison results are shown in Table~\ref{tab2}.
%Specifically, the online running time is composed of the query processing time and the gallery screening time and the ANN achieves acceleration in the screening process, thus we present the two time values in detail in Table~\ref{tab2}.
The Late2Early method is speeded up by using ANN during the screening process. 
%at the expense of serious damage to performance.
The proposed framework also achieves the screening acceleration over the Late2Early method, which is superior to ANN in most cases.
%by taking less screening time than the PQ-based Late2Early method and about the same screening time as the IVFPQ-based Late2Early method.
More importantly, the proposed framework is nearly the same as the Late2Early method on performance, while the ANN seriously hurts the performance of the Late2Early.

\begin{table*}[tb!]
\caption{Comparison of the proposed framework with various settings. The Adv. and Com. refer to the proposed advanced classification and the common one, respectively. ASL and BCE refer to the asymmetric loss and the binary cross-entropy loss, respectively. GT refers to the ground-truth annotations. NVA refers to the nouns, verbs and adjectives in the text used as the ground-truth annotations. For clarity, we present the R@(sum) variation of the proposed framework with other settings compared to the proposed framework (\emph{i.e.}, Ours).}
         \begin{center}
         \renewcommand\arraystretch{1.8}
\resizebox{1\textwidth}{!}{
         \begin{tabular}{l|c c | c c | c c | c c c | c c ||c c c c| c c c c}
         \Xhline{1.3pt}
         {\multirow{3}{*}{Method}} & \multicolumn{11}{c||}{Setting} & \multicolumn{4}{c|}{Flickr30K} & \multicolumn{4}{c}{MS-COCO} \\
         \cline{2-20}
          & \multicolumn{2}{c|}{Classifier} & \multicolumn{2}{c|}{Loss} & \multicolumn{2}{c|}{Image keyword} & \multicolumn{3}{c|}{Text keyword} & \multicolumn{2}{c||}{Keyword attribute} & \multicolumn{2}{c}{TR} & \multicolumn{2}{c|}{IR} & \multicolumn{2}{c}{TR} & \multicolumn{2}{c}{IR} \\
          & Adv. & Com. & ASL & BCE & Predict & GT & Predict & Extract & GT & Noun & NVA & R@1 & R@sum & R@1 & R@sum & R@1 & R@sum & R@1 & R@sum \\
          \hline
          \multicolumn{12}{l||}{Baseline\#1: ALBEF$_{0}$+ALBEF$_{all}$ (Late2Early method)} & 95.9 & 195.7 & 85.5 & 183.0 & 77.6 & 171.9 & 60.7 & 145.0 \\
          \cline{1-12}
          +Ours & \checkmark & $\times$ & \checkmark & $\times$ & \checkmark & $\times$ & \checkmark & $\times$ & $\times$ & \checkmark & $\times$ & 95.9 & 195.7 & 85.2 & 182.2 & 77.6 & 171.7 & 60.5 & 144.5 \\
          +Ours(Com.) & $\times$ & \checkmark & \checkmark & $\times$ & \checkmark & $\times$ & \checkmark & $\times$ & $\times$ & \checkmark & $\times$ & 95.8 & 195.6 (\textcolor[rgb]{1,0,0}{$\downarrow0.1$}) & 84.6 & 181.0 (\textcolor[rgb]{1,0,0}{$\downarrow1.2$}) & 77.5 & 171.4 (\textcolor[rgb]{1,0,0}{$\downarrow0.3$}) & 59.7 & 142.8 (\textcolor[rgb]{1,0,0}{$\downarrow1.7$})\\
          +Ours(BCE) & \checkmark & $\times$ & $\times$ & \checkmark & \checkmark & $\times$ & \checkmark & $\times$ & $\times$ & \checkmark & $\times$ & 95.8 & 195.6 (\textcolor[rgb]{1,0,0}{$\downarrow0.1$}) & 85.3 & 182.4 (\textcolor[rgb]{0,1,0}{$\uparrow0.2$}) & 77.7 & 171.8 (\textcolor[rgb]{0,1,0}{$\uparrow0.1$}) & 60.5 & 144.5 (\textcolor[rgb]{0,1,0}{$\rightarrow$}) \\
          +Ours(Ext.) & \checkmark & $\times$ & \checkmark & $\times$ & \checkmark & $\times$ & $\times$ & \checkmark & $\times$ & \checkmark & $\times$ & 95.6 & 195.3 (\textcolor[rgb]{1,0,0}{$\downarrow0.4$}) & 83.3 & 177.9 (\textcolor[rgb]{1,0,0}{$\downarrow4.3$}) & 77.7 & 171.8 (\textcolor[rgb]{0,1,0}{$\uparrow0.1$}) & 60.1 & 143.6 (\textcolor[rgb]{1,0,0}{$\downarrow0.9$}) \\
          +Ours(Merge) & \checkmark & $\times$ & \checkmark & $\times$ & \checkmark & $\times$ & \checkmark & \checkmark & $\times$ & \checkmark & $\times$ & 95.9 & 195.7 (\textcolor[rgb]{0,1,0}{$\rightarrow$}) & 85.3 & 182.2 (\textcolor[rgb]{0,1,0}{$\rightarrow$}) & 77.6 & 171.7 (\textcolor[rgb]{0,1,0}{$\rightarrow$}) & 60.6 & 144.8 (\textcolor[rgb]{0,1,0}{$\uparrow0.3$}) \\
          +Ours(NVA) & \checkmark & $\times$ & \checkmark & $\times$ & \checkmark & $\times$ & \checkmark& $\times$ & $\times$ & $\times$ & \checkmark & 95.8 & 195.5 (\textcolor[rgb]{1,0,0}{$\downarrow0.2$}) & 85.0 & 181.8 (\textcolor[rgb]{1,0,0}{$\downarrow0.4$}) & 77.7 & 171.8 (\textcolor[rgb]{0,1,0}{$\uparrow0.1$}) & 60.5 & 144.6 (\textcolor[rgb]{0,1,0}{$\uparrow0.1$}) \\
          +Ours(GT) & \checkmark & $\times$ & \checkmark & $\times$ & $\times$ & \checkmark & $\times$  & $\times$ & \checkmark & \checkmark & $\times$ & 96.1 & 196.0 & 86.4 & 184.3 & 77.8 & 172.1 & 61.1 & 145.8\\
          \hline
          \multicolumn{12}{l||}{Baseline\#2: ALBEF$_{0}$+ViLT-B/32a (Late2Early method)} & 85.6 & 182.8 & 66.7 & 156.8 & 65.6 & 155.1 & 46.0 & 121.8 \\
          \cline{1-12}
          +Ours & \checkmark & $\times$ & \checkmark & $\times$ & \checkmark & $\times$ & \checkmark & $\times$ & $\times$ & \checkmark & $\times$ & 86.4 & 184.8 & 66.2 & 156.4 & 65.6 & 155.0 & 46.1 & 121.4 \\
          +Ours(Com.) & $\times$ & \checkmark & \checkmark & $\times$ & \checkmark & $\times$ & \checkmark & $\times$ & $\times$ & \checkmark & $\times$ & 85.6 & 183.7 (\textcolor[rgb]{1,0,0}{$\downarrow1.1$}) & 66.5 & 156.4 (\textcolor[rgb]{0,1,0}{$\rightarrow$}) & 65.3 & 154.2 (\textcolor[rgb]{1,0,0}{$\downarrow0.8$}) & 45.5 & 119.3 (\textcolor[rgb]{1,0,0}{$\downarrow2.1$}) \\
          +Ours(BCE) & \checkmark & $\times$ & $\times$ & \checkmark & \checkmark & $\times$ & \checkmark & $\times$ & $\times$ & \checkmark & $\times$ & 86.0 & 184.2 (\textcolor[rgb]{1,0,0}{$\downarrow0.6$}) & 67.3 & 157.7 (\textcolor[rgb]{0,1,0}{$\uparrow1.3$}) & 65.5 & 154.8 (\textcolor[rgb]{1,0,0}{$\downarrow0.2$}) & 45.9 & 120.7 (\textcolor[rgb]{1,0,0}{$\downarrow0.7$}) \\
          +Ours(Ext.) & \checkmark & $\times$ & \checkmark & $\times$ & \checkmark & $\times$ & $\times$ & \checkmark & $\times$ & \checkmark & $\times$ & 85.3 & 183.0 (\textcolor[rgb]{1,0,0}{$\downarrow1.8$}) & 65.7 & 153.8 (\textcolor[rgb]{1,0,0}{$\downarrow2.6$}) & 65.7 & 155.3 (\textcolor[rgb]{0,1,0}{$\uparrow0.3$}) & 45.8 & 120.6 (\textcolor[rgb]{1,0,0}{$\downarrow0.8$}) \\
          +Ours(Merge) & \checkmark & $\times$ & \checkmark & $\times$ & \checkmark & $\times$ & \checkmark & \checkmark & $\times$ & \checkmark & $\times$ & 86.0 & 183.6 (\textcolor[rgb]{1,0,0}{$\downarrow1.2$}) & 66.9 & 157.2 (\textcolor[rgb]{0,1,0}{$\uparrow0.8$}) & 65.8 & 155.5 (\textcolor[rgb]{0,1,0}{$\downarrow0.5$}) & 45.8 & 120.9 (\textcolor[rgb]{1,0,0}{$\downarrow0.5$}) \\
          +Ours(NVA) & \checkmark & $\times$ & \checkmark & $\times$ & \checkmark & $\times$ & \checkmark& $\times$ & $\times$ & $\times$ & \checkmark & 85.9 & 184.1 (\textcolor[rgb]{1,0,0}{$\downarrow0.7$}) & 67.1 & 157.2 (\textcolor[rgb]{0,1,0}{$\uparrow0.8$}) & 65.7 & 155.5 (\textcolor[rgb]{0,1,0}{$\uparrow0.5$}) & 45.8 & 121.2 (\textcolor[rgb]{1,0,0}{$\downarrow0.2$}) \\
          +Ours(GT) & \checkmark & $\times$ & \checkmark & $\times$ & $\times$ & \checkmark & $\times$  & $\times$ & \checkmark & \checkmark & $\times$ & 87.0 & 185.1 & 69.7 & 161.8 & 65.7 & 154.5 & 46.9 & 122.9 \\
          \hline
          \multicolumn{12}{l||}{Baseline\#3: ViLT-B/32a (Early fusion method)} & 83.7 & 180.5 & 64.3 & 153.1 & 61.6 & 148.0 & 42.8 & 115.7 \\
          \cline{1-12}
          +Ours & \checkmark & $\times$ & \checkmark & $\times$ & \checkmark & $\times$ & \checkmark & $\times$ & $\times$ & \checkmark & $\times$ & 83.7 & 180.5 & 65.2 & 154.2 & 61.9 & 148.4 & 44.8 & 118.5 \\
          +Ours(Com.) & $\times$ & \checkmark & \checkmark & $\times$ & \checkmark & $\times$ & \checkmark & $\times$ & $\times$ & \checkmark & $\times$ & 83.5 & 180.0 (\textcolor[rgb]{1,0,0}{$\downarrow0.5$}) & 65.6 & 154.2 (\textcolor[rgb]{0,1,0}{$\rightarrow$}) & 61.7 & 148.1 (\textcolor[rgb]{1,0,0}{$\downarrow0.3$}) & 44.5 & 117.2 (\textcolor[rgb]{1,0,0}{$\downarrow1.3$}) \\
          +Ours(BCE) & \checkmark & $\times$ & $\times$ & \checkmark & \checkmark & $\times$ & \checkmark & $\times$ & $\times$ & \checkmark & $\times$ & 83.7 & 180.7 (\textcolor[rgb]{0,1,0}{$\uparrow0.2$}) & 66.2 & 154.6 (\textcolor[rgb]{0,1,0}{$\uparrow0.4$}) & 61.9 & 148.4 (\textcolor[rgb]{0,1,0}{$\rightarrow$}) & 44.9 & 118.5 (\textcolor[rgb]{0,1,0}{$\rightarrow$}) \\
          +Ours(Ext.) & \checkmark & $\times$ & \checkmark & $\times$ & \checkmark & $\times$ & $\times$ & \checkmark & $\times$ & \checkmark & $\times$ & 83.3 & 179.7 (\textcolor[rgb]{1,0,0}{$\downarrow0.8$}) & 64.7 & 151.5 (\textcolor[rgb]{1,0,0}{$\downarrow2.7$}) & 62.2 & 148.7 (\textcolor[rgb]{0,1,0}{$\uparrow0.3$}) & 44.8 & 117.8 (\textcolor[rgb]{1,0,0}{$\downarrow0.7$}) \\
          +Ours(Merge) & \checkmark & $\times$ & \checkmark & $\times$ & \checkmark & $\times$ & \checkmark & \checkmark & $\times$ & \checkmark & $\times$ & 83.8 & 180.6 (\textcolor[rgb]{0,1,0}{$\uparrow0.1$}) & 65.7 & 155.0 (\textcolor[rgb]{0,1,0}{$\uparrow0.8$}) & 61.9 & 148.4 (\textcolor[rgb]{0,1,0}{$\rightarrow$}) & 44.8 & 118.4 (\textcolor[rgb]{0,1,0}{$\uparrow0.1$}) \\
          +Ours(NVA) & \checkmark & $\times$ & \checkmark & $\times$ & \checkmark & $\times$ & \checkmark& $\times$ & $\times$ & $\times$ & \checkmark & 83.5 & 180.0 (\textcolor[rgb]{1,0,0}{$\downarrow0.5$}) & 66.5 & 155.6 (\textcolor[rgb]{0,1,0}{$\uparrow1.4$}) & 61.9 & 148.4 (\textcolor[rgb]{0,1,0}{$\rightarrow$}) & 44.7 & 118.4 (\textcolor[rgb]{1,0,0}{$\downarrow0.1$}) \\
          +Ours(GT) & \checkmark & $\times$ & \checkmark & $\times$ & $\times$ & \checkmark & $\times$  & $\times$ & \checkmark & \checkmark & $\times$ & 85.6 & 183.1 & 68.1 & 159.7 & 62.6 & 149.8 & 46.1 & 120.9 \\
          \Xhline{1.2pt}
         \end{tabular}}   
         \label{tab3}
         \end{center}
\end{table*}

\textbf{Comparison with common classification.}
The performance of the multi-label classification for keyword prediction directly influences the result of the proposed framework.
Aiming at a high-performance classification, we propose the advanced classification.
Alternatively, the classification can be replaced with any existing common one.
For comparison, we use the ViT-B/16~\cite{dosovitskiy2020image} and BERT~\cite{devlin2018bert} together with the ASL as the image and text classifiers, respectively. 
In terms of performance, as shown in Table~\ref{tab3}, the proposed framework with the advanced classification (Ours) has a consistent advantage over that with the common classification (Ours(Com.)) based on different baselines, which verifies the superiority of the proposed advanced classification.
In terms of the classification-related metrics, we show the number of model parameters, the recall rate of ground-truth gallery samples and the mAP values of image and text classifiers in Table~\ref{tab4}.
The advanced classification model surpasses the common one at all metrics.
Specifically, the number of parameters for the proposed framework with the advanced classification is far less than that with the common classification, which shows the superiority of the proposed framework in resource consumption.

\textbf{Analysis of the advanced classification.}
We analyze the proposed advanced classification from three aspects.
(1) We adopt the common binary cross-entropy loss (BCE) to replace the ASL as the classification loss.
The comparison results (Ours vs. Ours(BCE)) in Table~\ref{tab3} show that the `Ours(BCE)' has a slight drop at R@sum compared to the `Ours' on Flickr30K (TR) when applied in the Baseline\#1 and Baseline\#2 and on MS-COCO when applied in Baseline\#2, and has a slight increase or remains unchanged in other cases.
In general, the `Ours' and `Ours(BCE)' are much of a muchness on performance.
By adopting other state-of-the-art classification losses, the proposed framework with high flexibility is expected to further improve.
(2) We develop a multi-label text classifier for text keyword prediction.
Instead, we can also extract the nouns directly from the text by using the natural language toolkit (Ours(Ext.)), or use both the predicted labels and extracted nouns as the keywords (Ours(Merge)). 
As shown in Table~\ref{tab3}, the `Ours' performs better than the `Ours(Ext.)' in all cases except for MS-COCO (TR).
In the `Ours(Ext.)', the text keywords are derived from itself.
In contrast, the keywords are from all ground-truth annotations in the `Ours', enabling the exploration of keyword synonyms and positively affecting performance.
However, the classifier is likely to predict the wrong keywords in the `Ours', thus hurting retrieval performance.
Beyond that, as mentioned in Section~\ref{ICKE}, there are pre-screening failures in the `Ours(Ext.)', yet not in the `Ours'. 
Specifically, $37$ and $9$ texts can not incorporate into the pre-screening process by the `Ours(Ext.)' in Flickr30K and MS-COCO, respectively.
In view of the above, merging the predicted labels and extracted nouns to the keywords is a complementary solution.
It can be seen from Table~\ref{tab3} that `Ours(Merge)' surpasses the `Ours' by a small margin in most cases.
At the same time, though, the `Ours(Merge)' suffers from a lower screening rate due to the increasing number of text keywords from the mergence operation.
As a result, the `Ours(Merge)' is weaker than the `Ours' on retrieval efficiency.
Specifically, the $2.0\times$ and $3.9\times$ speedups are reached by the `Ours(Merge)' on Flickr30K and MS-COCO, respectively; by contrast, the `Ours' introduces the $2.1\times$ and $4.1\times$ speedups.
(3) We use the nouns in the text as the image and text ground-truth annotations for training classifiers.
One alternative is to use the nouns, verbs and adjectives (Ours(NVA)).
The results in Table~\ref{tab3} show that the `Ours(NVA)' is inferior to the `Ours' on performance in some cases and performs better than the `Ours' in others. 
In general, the ‘Ours’ and ‘Ours(NVA)’ are much of a muchness on performance.
However, it is remarkable that more keywords can be predicted in the `Ours(NVA)' due to more diverse ground-truth annotations, as a result of which the screening rate is reduced and the retrieval efficiency is hurt.

\textbf{The upper bound of the proposed framework.}
%加retrieval结果
Excluding the harmful effects of the wrong classification results on performance, we directly use the ground-truth annotations as the keywords to explore the upper bound of the proposed framework on performance (Ours(GT)).
As shown in Table~\ref{tab3}, the `Ours(GT)' performs better than the `Ours' and also consistently achieves remarkable improvement over the baseline on performance.
Owing the high flexibility, the proposed framework is hopeful to realize a powerful win-win situation for accuracy and efficiency by adopting a better classification technique. 

\begin{table}[tb!]
\caption{Comparison of different classification techniques. The mAP(image) and mAP(text) refer to the mAP value of the image and text classifiers, respectively. The Adv. and Com. refer to the proposed advanced classification and the common one, respectively.}
         \begin{center}
         \renewcommand\arraystretch{1.8}
\resizebox{0.49\textwidth}{!}{
         \begin{tabular}{l||c c c|c c c| c}
         \Xhline{1.3pt}
         {\multirow{2}{*}{Method}} & \multicolumn{3}{c|}{Flickr30K} & \multicolumn{3}{c|}{MS-COCO} & {\multirow{2}{*}{Para(M)}} \\
          & Recall & mAP(image) & mAP(text) & Recall & mAP(image) & mAP(text) & \\
          \hline
          Adv. & 99.4 & 52.0 & 58.9 & 99.4 & 56.9 & 62.2 & 1.3 \\
          Com. & 98.7 & 48.4 & 58.9 & 98.1 & 50.5 & 61.9 & 196.4 \\
          \Xhline{1.2pt}
         \end{tabular}}  
         \label{tab4}
         \end{center}
\end{table}

\textbf{Parameter analysis.}
There are two parameters in the proposed framework, \emph{i.e.}, $R_I$ in image keyword prediction and $R_T$ in text keyword prediction.
%A higher value of $R_I$ or $R_T$ indicates that more gallery samples are kept in the pre-screening stage and enter into the follow-up retrieval network.
We show the results of the proposed framework with different parameters based on the ALBEF$_{0}$+ALBEF$_{all}$ in Table~\ref{tab5}.
With an increasing value of $R_I$ or $R_T$, more ground-truth gallery samples are kept in the pre-screening stage and enter into the follow-up retrieval network (\emph{i.e.}, a higher Recall value), resulting in better accuracy (\emph{i.e.}, a higher R@1), yet at the same time, causing a decline in efficiency (\emph{i.e.}, a lower Speedup).
For achieving a trade-off between accuracy and efficiency, we set $R_I=15$ and $R_T=3$ in the experiments.

\begin{table}[tb!]
\caption{Results of the proposed framework with different parameters. The ALBEF$_{0}$+ALBEF$_{all}$ is used as the baseline.}
         \begin{center}
         \renewcommand\arraystretch{1.8}
\resizebox{0.49\textwidth}{!}{
         \begin{tabular}{c||c c c c | c c c c}
         \Xhline{1.3pt}
         {\multirow{2}{*}{Para.}} & \multicolumn{4}{c|}{Flickr30K} & \multicolumn{4}{c}{MS-COCO} \\
         \cline{2-9}
          & R@1(TR) & R@1(IR) & Recall & Speedup & R@1(TR) & R@1(IR) & Recall & Speedup \\
          \hline
          \multicolumn{9}{l}{Varying $R_I$ and fixing $R_T=3$.} \\
          \hline
          5 & 95.5 & 83.0 & 96.4 & $\times$2.9 & 77.4 & 59.4 & 97.2 & $\times$7.0 \\
          10 & 95.8 & 84.6 & 98.6 & $\times$2.3 & 77.6 & 60.3 & 98.9 & $\times$5.0 \\
          15 & 95.9 & 85.2 & 99.4 & $\times$2.1 & 77.6 & 60.5 & 99.4 & $\times$4.1 \\
          20 & 95.9 & 85.2 & 99.6 & $\times$1.9 & 77.6 & 60.6 & 99.6 & $\times$3.7 \\
          25 & 95.9 & 85.3 & 99.8 & $\times$1.8 & 77.6 & 60.6 & 99.7 & $\times$3.3 \\
          30 & 95.9 & 85.3 & 99.8 & $\times$1.7 & 77.6 & 60.6 & 99.8 & $\times$3.1 \\
          \hline
          \multicolumn{9}{l}{Varying $R_T$ and fixing $R_I=15$.} \\ 
          \hline
          1 & 95.0 & 79.5 & 91.7 & $\times$3.3 & 76.9 & 57.1 & 92.4 & $\times$8.4 \\
          2 & 95.7 & 84.3 & 98.1 & $\times$2.5 & 77.5 & 60.0 & 98.3 & $\times$5.4 \\
          3 & 95.9 & 85.2 & 99.4 & $\times$2.1 & 77.6 & 60.5 & 99.4 & $\times$4.1 \\
          4 & 95.9 & 85.4 & 99.6 & $\times$1.8 & 77.7 & 60.6 & 99.7 & $\times$3.3 \\
          5 & 95.9 & 85.5 & 99.8 & $\times$1.7 & 77.7 & 60.6 & 99.8 & $\times$2.9 \\
          10 & 95.9 & 85.5 & 100.0 & $\times$1.3 & 77.7 & 60.7 & 100.0 & $\times$1.9 \\
          20 & 95.9 & 85.5 & 100.0 & $\times$1.1 & 77.7 & 60.7 & 100.0 & $\times$1.4 \\
          \Xhline{1.2pt}
         \end{tabular}}      
         \label{tab5}
         \end{center}
\end{table}

\begin{figure}[t]
\begin{center}
   \includegraphics[width=1\linewidth, height=0.43\linewidth]{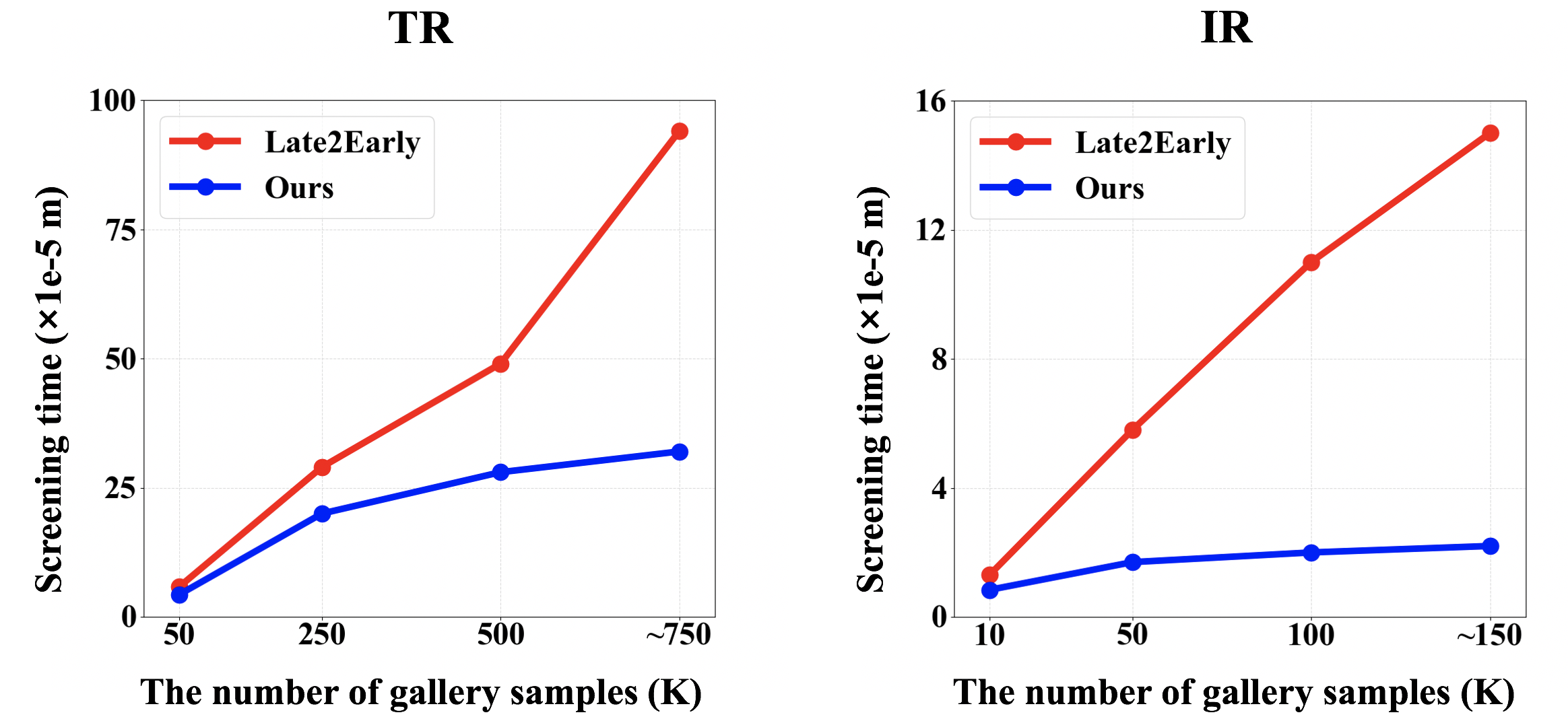}
\end{center}
   \caption{Running time (in minutes) of one query on large gallery data (in thousands). `$\sim$750' and `$\sim$150' specifically refer to 771,837 texts and 154,301 images, respectively.}
\label{fig6}
\end{figure}

\textbf{Efficiency comparison on larger-scale data.}
To verify the superiority of the proposed framework in reality, we make an efficiency comparison with the Late2Early method ALBEF$_{0}$+ALBEF$_{all}$ on larger-scale data, which is constructed by merging all data in Flickr30K and MS-COCO and results in $154,301$ images and $771,837$ texts.
We adopt the classifiers trained on MS-COCO in the proposed framework in the experiments.
Fig.~\ref{fig6} shows the running time\footnote{The running time is composed of the query processing time and gallery screening time and the change in the number of gallery samples only impacts the screening time, and thus we only report the screening time in Fig.~\ref{fig6}. The query processing time is 3.0e-2 in TR and 3.1e-4 in IR for the Late2Early method, and is 0 for the proposed framework thanks to the proposed multi-task learning scheme.} of one query on the new benchmark with different numbers of gallery samples.
The proposed framework shows significant superiority over the Late2Early method on efficiency.
Moreover, with an increasing number of the gallery samples, the increasing trend of the proposed framework's running time is gradually slowed down, by contrast, the Late2Early method's running time increases linearly.
These results indicate the potential application of the proposed framework in reality.

\begin{table}[tb!]
\caption{Performance of the proposed framework on Flickr30K-all.}
         \begin{center}
         \renewcommand\arraystretch{1.8}
\resizebox{0.49\textwidth}{!}{
         \begin{tabular}{l|l||c c c c c c c}
         \Xhline{1.3pt}
         {\multirow{2}{*}{Genre}} & {\multirow{2}{*}{Method}} & \multicolumn{3}{c}{TR} & \multicolumn{3}{c}{IR} & {\multirow{2}{*}{Speedup}} \\
         & & R@1 & R@5 & R@sum & R@1 & R@5 & R@sum &  \\
         \hline
         Late2Early & ALBEF$_{0}$+ALBEF$_{all}$ & 68.8 & 87.1 & 155.9 & 48.1 & 70.6 & 118.7 & ~~-  \\      
          & +Ours & 68.8 & 87.0 & 155.8 (\textcolor[rgb]{1,0,0}{$\downarrow0.1$}) & 47.8 & 70.2 & 118.0 (\textcolor[rgb]{1,0,0}{$\downarrow0.7$}) & 2.0$\times$ \\  
          \Xhline{1.2pt}
         \end{tabular}}
         \label{tab6}
         \end{center}
\end{table}

\textbf{Simulation of practical application scenario.}
In practical application scenario, the proposed framework usually deals with millions of wide-open inference data with more diverse content.
As a result, the keywords predefined from training data may not have appropriate coverage of the semantic space of the inference data, which tends to lead to false predictions for keywords and affect the framework’s performance.
To investigate the performance of the proposed framework in such a scenario, we perform the proposed framework on Flickr30K-all, which is constructed by merging the training, validataion and inference data in Flickr30K and results in $31,014$ images and $155,070$ texts and we adopt the proposed framework trained in MS-COCO.
There is a relatively big gap between the training and inference data.
Table~\ref{tab6} presents the results on Flickr30K-all.
It can be seen that a decent performance in accuracy and speedup can still be obtained by the proposed framework.
The proposed framework would work as long as there is only one correctly predicted keyword among the resulting keywords, so that the negative impact of the data gap on performance is alleviated to some extent.

\begin{figure*}[t]
\begin{center}
   \includegraphics[width=1\linewidth, height=0.45\linewidth]{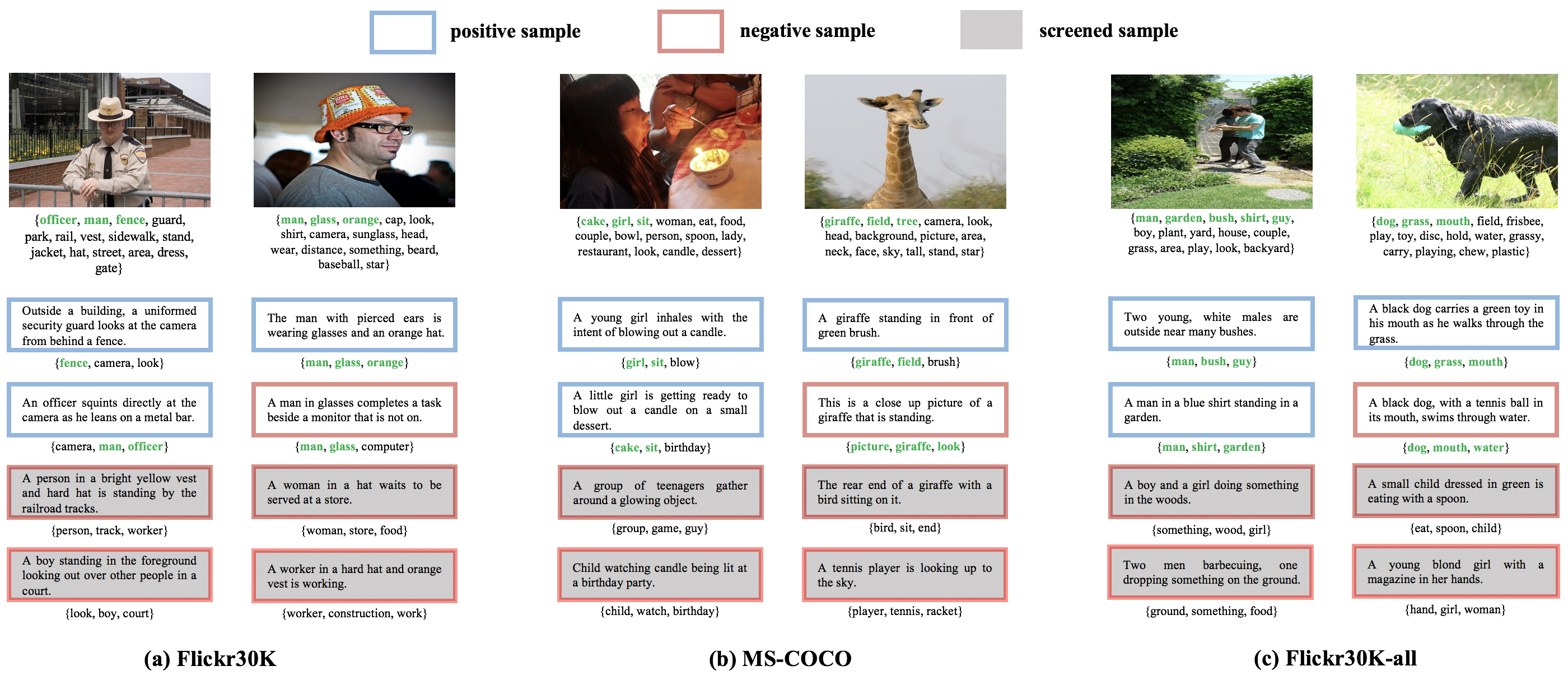}
\end{center}
   \caption{Visualizing the predicted keywords and the screening results from the proposed framework taking TR example. The keywords enclosed by the curly braces `$\{\}$' are below the sample. We randomly select a few samples among large number of gallery samples for visualization.}
\label{fig7}
\end{figure*}

\textbf{Visualization and limitation.}
We visualize the predicted keywords and the screening results\footnote{For lack of space, we randomly select a few samples among large number of gallery samples for visualization.} from the proposed framework in Fig.~\ref{fig7}.
It can be seen that the proposed framework yields the correct and reasonable keywords at the semantic level and can bring decent screening results, \emph{i.e.,} screening out the gallery samples semantically irrelevant to the query.
Notably, the results in Flickr30K-all are obtained by the framework trained in MS-COCO and the gap between the training and inference data does not have much effect on performance.

Taking a step further, we discuss the limitation of the proposed framework.
The proposed framework relies on the quality of the predicted keywords.
With that in mind, we propose a multi-task learning scheme for improving the classification accuracy and prompting the keyword prediction, and present the detailed results of the classifier with various settings in Table~\ref{tab3} and the in-depth analysis in Section~\ref{FA}.
Nevertheless, there still leaves room for further development of the proposed framework with a more powerful classification technique. 
We could consider the labels from the object detection as a powerful supplementary to the classification results, which meanwhile could well apply to the case of the collections without extensive textual annotation.

\section{Conclusion}

This paper focuses on the low efficiency problem in the image-text retrieval task.
Admittedly, the existing image-text retrieval methods suffer from at least $O(N)$ time complexity and may not be economically practical in many real cases.
%Existing image-text retrieval methods suffer from N-related time complexity and it is not economically practical to apply it in reality.
To this end, we present a simple and effective keyword-guided pre-screening framework, in which the image and text samples are projected into the keywords, and then a fast keyword matching across modalities is executed to screen out the gallery samples irrelevant to the query sample.
The remaining gallery samples with an amount that is much less than the number of the original gallery samples are fed into the common image-text retrieval network, thus realizing the retrieval acceleration.
%In particular, we project the samples into the keywords by a multi-label classification and propose a multi-task learning scheme on the image-text retrieval network for a powerful classification correspondingly; we introduce the inverted index technique into the keyword matching process, greatly accelerating the pre-screening.
The proposed framework is characterized by low consumption and excellent compatibility.
We experimentally verify the effectiveness of the proposed framework. 
%We realize the win-win situation between accuracy and efficiency by the proposed framework.
%In future work, we will focus on optimizing the keyword selection for better pre-screening, and we can consider the hyponymy relations and the synonyms of the predicted nouns.

\ifCLASSOPTIONcaptionsoff
  \newpage
\fi



\bibliographystyle{IEEEtran}
\bibliography{sample-base}



%\vfill


\end{document}